\documentclass{article}
\usepackage{iclr2027_conference,times}
\iclrfinalcopy
\usepackage[T1]{fontenc}
\usepackage[utf8]{inputenc}
\usepackage{microtype,amsmath,amssymb,amsthm,booktabs,array,tabularx,multirow,graphicx}
\usepackage{algorithm,algpseudocode}
\usepackage{xcolor,url,hyperref,placeins}
\usepackage{longtable}
\usepackage{needspace}
\usepackage[shortlabels]{enumitem}
\hypersetup{hidelinks,pdftitle={Does the Workspace Do the Work? A Causal Protocol for Verifying Global-Workspace Computation in a Language Model},pdfauthor={Diaoxuan (Nora) He, Linqian Wu},pdfsubject={Causal verification, Global Workspace Theory, and model organisms}}
\author{Diaoxuan (Nora) He\\University of Debrecen \And Linqian Wu\thanks{Corresponding author.}\\University of Waterloo\\Perimeter Institute for Theoretical Physics\\\texttt{lwu1@perimeterinstitute.ca}}

\newcommand{\R}{\mathbb{R}}

\newcommand{\NULL}{\textsc{Null}}

\newcolumntype{L}[1]{>{\raggedright\arraybackslash}p{#1}}
\newtheorem{proposition}{Proposition}
\newtheorem{corollary}{Corollary}
\newtheorem{definition}{Definition}
\newcommand{\rev}[1]{#1}
\makeatletter\let\inputrows\@@input\makeatother
\title{Does the Workspace Do the Work? A Causal Protocol for Verifying Global-Workspace Computation in a Language Model}
\date{}
\begin{document}
\maketitle
\lhead{}
\rhead{}
\renewcommand{\headrulewidth}{0pt}
\begin{abstract}
Global Workspace Theory (GWT), a leading theory of consciousness, holds that information becomes conscious when it is broadcast to specialized processes that use it. Building a workspace into a network does not reveal what the trained network computes with it: a module that reports an answer may have combined a broadcast message with its own knowledge, or merely forwarded an answer computed elsewhere, and task accuracy and ablation cannot tell these apart. We introduce a causal protocol that identifies who computes when modules communicate. Its central test transplants a message from another input, chosen so that ignoring, relaying and recombining it predict three different answers. On networks built to relay or to recombine, the test assigns the right computation in every case. Among trained workspace models it finds all three outcomes: one leaves its broadcast nearly unused, and of two models that differ only in their training data, one relays and the other recombines. In AFGW, a language model whose modules communicate only through a learned workspace, a transplanted message yields the answer that recombination predicts in 97\% of cases, and its content acts by redirecting what the receiver retrieves from its own memory. The tests carry over to other communication designs and to a known interface of a pretrained Transformer, turning GWT's central claim into a testable property of trained models.
\end{abstract}
\newlength{\vthirtyninestextfloatsep}
\newlength{\vthirtyninesfloatsep}
\newlength{\vthirtyninesintextsep}
\setlength{\vthirtyninestextfloatsep}{\textfloatsep}
\setlength{\vthirtyninesfloatsep}{\floatsep}
\setlength{\vthirtyninesintextsep}{\intextsep}
\setlength{\textfloatsep}{12pt plus 2pt minus 3pt}
\setlength{\floatsep}{10pt plus 2pt minus 2pt}
\setlength{\intextsep}{6pt plus 2pt minus 2pt}
\section{Introduction}

\label{sec:introduction}

When the modules of a network communicate, which of them does the computing? The question sits at the heart of Global Workspace Theory (GWT; \citealp{baars1988,dehaene1998,dehaene2011,mashour2020}), a leading scientific theory of consciousness. GWT holds that information becomes consciously accessible when it enters a limited-capacity workspace and is broadcast to specialized processes that \emph{use} it \citep{block1995}. On this view, what makes information conscious is not where it is stored but whether other processes can access and use it, so the theory's central claim is a claim about use. Indicator-based assessments check models for workspace properties \citep{butlin2023,chalmers2023}, interpretability work reports workspace-like representations in pretrained language models \citep{gurnee2026}, and networks with trained workspaces serve as experimental substrates for the theory \citep{goyal2022,albantakis2025}. All three raise one question: does a system that has a workspace perform the computation the theory assigns to it?

\begin{figure}[t]
\centering\includegraphics[width=\linewidth]{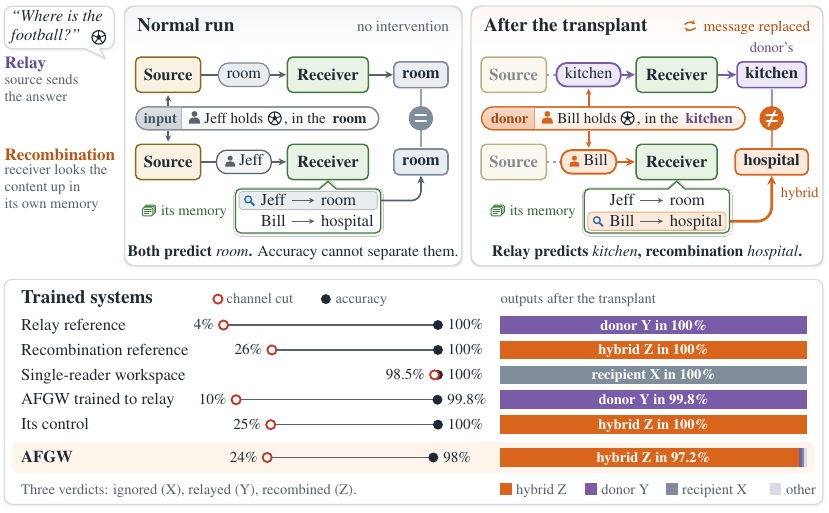}
\caption{\textbf{One swapped message separates relay from recombination, which accuracy and ablation cannot.} Top: the same question answered by relay and by recombination, before and after the donor's message is swapped in. Bottom: accuracy with the channel intact and cut, and the outputs after the transplant, for trained reference systems of each kind, \rev{a single-reader workspace model that ignores its broadcast, an AFGW model trained to relay with its control, and the four main AFGW models.} $X$, $Y$ and $Z$ are the answers that ignoring, relaying and recombining the message predict.}
\label{fig:teaser}
\end{figure}

This is rarely tested, and task accuracy and ablation cannot test it. Consider two modules that exchange messages (Figure~\ref{fig:teaser}, top). The source knows who holds each object, and the receiver knows where each person is. Asked ``Where is the football?'', the network answers correctly, and it fails when the channel is cut. These facts fit two opposite computations. The source may send the holder's name, which the receiver looks up in its own memory. This is \emph{recombination}, the computation GWT describes. Or the message may already carry the final answer, which the receiver merely forwards. This is \emph{relay}. Ablation shows only that information flows through the channel. Swapping in a message from another input, the standard patch, does not settle the matter either when, as is usual, the forwarded and the recombined answer are the same string. We call the distance between building a channel and knowing what a trained network computes with it the \emph{verification gap}.

We close this gap with a causal protocol that identifies how communicating modules divide computational work. Its central test chooses the swapped-in message so that the candidate computations predict different answers. In the recipient input, Jeff holds the football and is in the room, while Bill is in the hospital. In the donor input, Bill holds the football and is in the kitchen. After the donor's message is transplanted, ignoring it predicts \emph{room}, relaying the donor's answer predicts \emph{kitchen}, and combining the message with the receiver's own memory predicts \emph{hospital}. We call this third answer the \emph{hybrid answer}. Three further tests ask whether the model depends on the channel, how the content acts inside the receiver, and whether one message serves many receivers (Figure~\ref{fig:ladder}). \rev{The protocol builds on established interventions \citep{zhang2024,geiger2024,wang2023,variengien2025,prakash2026} and adds three things: a design that gives each competing account of communication its own prediction, a calibration on systems whose computation is known, and a test of how communicated content is consumed.} We make three contributions.

\begin{itemize}[leftmargin=*,itemsep=1pt,topsep=2pt,parsep=0pt]

\item \textbf{A calibrated protocol that identifies who computes.} Each of its four tests pairs an intervention with the alternative it rules out (Section~\ref{sec:protocol}, Table~\ref{tab:rungs}). \rev{On reference systems built to relay or to recombine, it assigns the right computation in every one of 4,096 cases. On trained workspace models it returns all three verdicts: a single-reader model ignores its broadcast, a second model relays and a third recombines. The last two differ only in their training data, and accuracy, ablation and a standard patch give the same result for both.}

\item \textbf{A verified mechanism in a learned workspace.} AFGW is a 6.9M-parameter language model whose modules exchange information only through a learned workspace, with nothing supervised except answer tokens. \rev{In four models that answer 98\% of held-out composition questions, one transplanted message yields the hybrid answer in 97\% of cases, against 0.4\% without intervention.} The content acts by redirecting retrieval: restoring only the receiver's memory queries undoes the transplant in 2,047 of 2,048 cases, and installing only the queries the transplant induces reproduces it in \rev{2,047} (Section~\ref{sec:results}).

\item \rev{\textbf{Evidence that the tests travel.} These results hold in worlds twice as large, across communication and memory-reading designs, and at a known interface of a pretrained Transformer. When a second route makes the workspace replaceable, zero-ablation calls it nearly unnecessary, yet the transplant still detects its use (Section~\ref{sec:beyond}, Figure~\ref{fig:generality}).}

\end{itemize}

The results serve two fields. For the science of consciousness, they turn GWT's central functional claim into a property that can be measured, and refuted, in trained systems. For interpretability, they supply a test of who computes wherever modules, experts, retrievers or agents pass messages, and a model organism with exact counterfactual ground truth on which such methods can be calibrated.

\section{AFGW: A Model Organism Built for Verification}

\label{sec:architecture}

In a large pretrained model, an intervention on a candidate workspace leaves other pathways open. We therefore first apply the protocol to a model organism \citep{elhage2022,nanda2023} in which every route can be enumerated. AFGW (Access-First Global Workspace) is built around access, the availability of broadcast content to other modules. It builds in three commitments of GWT, specialized processors, competition for limited shared access and global broadcast, and leaves the fourth, use of that content by receivers, to the protocol (Figure~\ref{fig:architecture}; Table~\ref{tab:commitments} in Appendix~\ref{app:architecture}).

\begin{figure}[H]
\centering\includegraphics[width=.88\linewidth]{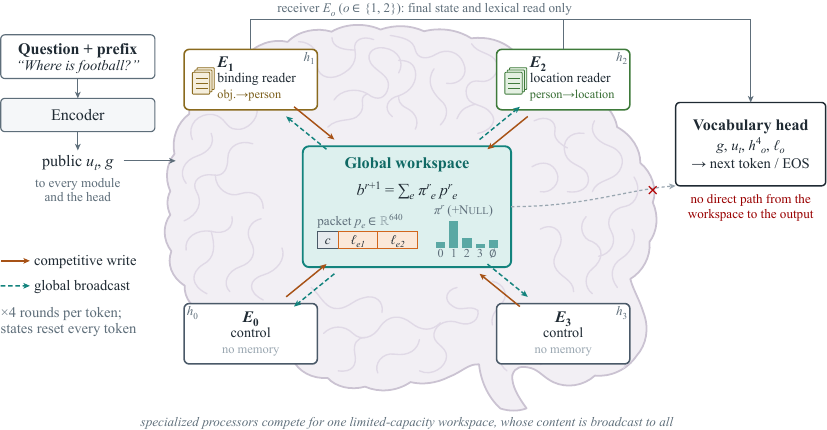}
\caption{\textbf{AFGW computes each generated token in four rounds.} In each round, every module writes a packet. The packets compete with a NULL option, and the pooled message is broadcast to all modules before their next read. A packet holds a control block $c$ and its lexical reads $\ell$.}
\label{fig:architecture}
\end{figure}

\textbf{Task.} Each input describes a small world of four objects, four people, and four locations through two kinds of fact records. Binding records map objects to people (``The holder of football is Jeff.''). Location records map people to locations (``Jeff is in the room.''). Local questions query one relation. Composition questions need both: ``Where is the football?'' requires the holder of the football and then that person's location. Reverse questions map a location back to an object.

\textbf{Modules.} AFGW has four recurrent modules. The binding reader $E_1$ reads only binding records, the location reader $E_2$ reads only location records, and $E_0$ and $E_3$ have no memory and only compete for the workspace. Records are encoded independently, so composition requires information to cross between modules. The question type determines the answering module, the \emph{receiver} $o$ ($E_2$ for ``Where is the football?''). The other fact reader, which holds the fact the receiver lacks ($E_1$ here), is the \emph{source}.

\textbf{Rounds.} Each generated token is computed in four rounds, $r = 0,\ldots,3$, starting from zero private states $h_e$ and an empty workspace. In each round, a fact-reading module $e$ queries its own memory $\mathcal{B}_e$. The queries depend on its private state, on features that every module sees ($g$ of the question, $u_t$ of the question and the answer so far), and on the current broadcast $b^r$:

\begin{equation}
q_e^r = Q_e(h_e^r,\,b^r,\,u_t,\,g), \qquad \ell_e^r=\mathrm{Read}(q_e^r,\mathcal{B}_e).
\label{eq:query}\end{equation}

The read returns a contextual summary of the selected records and the \emph{lexical read} $\ell_e^r$, a weighted sum of token embeddings that the output head can emit as words. The module updates its state from the contextual summary and the broadcast with a gated recurrent unit \citep{cho2014} and writes a packet $p_e^r$, a control block computed from its reads plus the lexical reads themselves, which we call write $r$. The packets compete for the workspace, and the broadcast is their weighted mixture

\begin{equation}
b^{r+1}=\textstyle\sum_e \pi_e^r p_e^r,
\label{eq:broadcast}\end{equation}

where $\pi^r$ is a learned softmax over the four modules and an empty NULL option. Every module receives the same broadcast, and it enters every reader's query in the next round. Write $r$ is thus read in round $r+1$: write 2 enters the broadcast $b^3$ and shapes the receiver's round-3 read, its last before it answers. One module's knowledge can therefore change what another module retrieves next. All states reset at every token, and the last round's packet is never read, a built-in negative control.

\textbf{Output and identifiability.} The vocabulary head sees only $u_t$ and $g$, the receiver's final state $h_o^4$ (its state after the four rounds), and the receiver's final lexical read. It has no direct input from the workspace, so broadcast content can matter only through what a receiver does with it. \rev{Without this boundary and the split memory, a trained model can bypass its broadcast (Section~\ref{sec:whocomputes}).} Decoding is greedy and ends with an end-of-sequence token (EOS). Everything except memory access and the choice of answering module is learned from final-answer tokens (Appendix~\ref{app:architecture}). Two further properties make the tests exact. The workspace is the only route between modules, so switching it off has a provable information bound (Proposition~\ref{prop:off}). And the broadcast reaches a receiver by two routes, through its memory queries and directly through its private state, so which route carries the content is left to learning.

\section{The Protocol: From Presence to Use}

\label{sec:protocol}

\begin{figure}[H]
\centering\includegraphics[width=\linewidth]{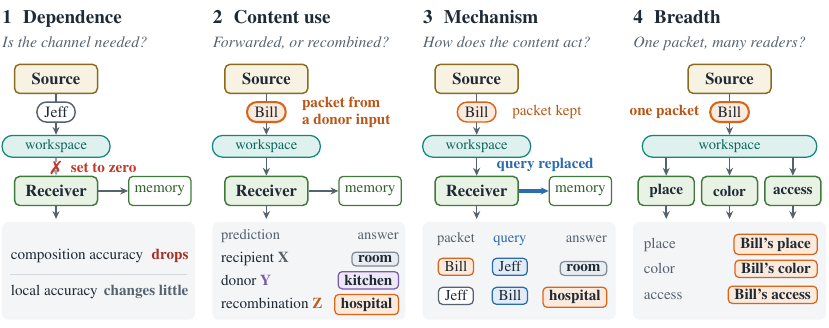}
\caption{\textbf{Four tests, from dependence to breadth.} The same schematic locates each intervention: zeroing every received broadcast, replacing the source's packet with its packet from a donor input, restoring or installing the receiver's queries, and delivering one packet to three receivers (Table~\ref{tab:rungs}).}
\label{fig:ladder}
\end{figure}

Building a broadcast channel establishes only that it is present. The protocol asks four further questions in order, each answered by one intervention that rules out the alternative left open by the previous one (Figure~\ref{fig:ladder}, Table~\ref{tab:rungs}). We call the four tests rungs: rung 2 is the \emph{content test}, and the two interventions of rung 3 are the \emph{read tests}. Table~\ref{tab:terms} defines the terms on a running example.

\begin{table}[t]
\caption{\rev{Terms used throughout, illustrated with the running example (``Where is the football?'').}}
\label{tab:terms}
\centering\scriptsize
\setlength{\tabcolsep}{3pt}
\begin{tabularx}{\linewidth}{@{}>{\raggedright\arraybackslash}p{.16\linewidth}>{\raggedright\arraybackslash}X>{\raggedright\arraybackslash}p{.25\linewidth}@{}}
\toprule
Term & Meaning & In the example\\
\midrule
recipient, donor & the input on which the model is run and intervened on, and the input from which a packet is taken & Jeff holds the football. Bill does.\\
receiver, source & the module that answers the question, and the other fact reader, which holds the fact the receiver lacks & $E_2$ (locations), $E_1$ (bindings)\\
$X$, $Y$, $Z$ & the correct answers in the recipient world, the donor world, and the hybrid world, which takes the source's facts from the donor and the receiver's facts from the recipient (the primary evaluation sets use only cases where all three differ) & room, kitchen, hospital. The hybrid world has the donor's bindings and the recipient's locations.\\
native, transplant & a run with no intervention, and the same run with a packet of the source replaced by its packet from the donor's native run (write 2 unless stated) & answers room. $E_1$'s packet names Bill.\\
relay, recombination & the receiver passes on the donor's answer ($Y$), or it combines the transplanted content with its own memory ($Z$) & kitchen, hospital\\
\bottomrule
\end{tabularx}
\end{table}

\subsection{Rung 1: is the channel needed?}

\label{sec:off}

\rev{The first test switches the workspace off and compares composition questions, which need both memories, with local questions, which need one.

\begin{definition}[OFF]\label{def:off}
OFF sets to zero every broadcast $b^r$ received by every module, in every round $r=1,2,3$ ($b^0$ is already zero) and at every generated token including EOS. Memories, reads, state updates, packet writes, routing, and the output head are unchanged.
\end{definition}

\begin{samepage}
\begin{proposition}[OFF information bound]\label{prop:off}
Under OFF, the receiver's complete generated answer equals $F_\theta(Q,\mathcal{B}_o)$, a deterministic function of the question $Q$ and the records $\mathcal{B}_o$ in the receiver's own memory. Inputs that agree on $Q$ and $\mathcal{B}_o$ therefore receive the same answer, and OFF accuracy is at most that of the best predictor that sees only $(Q,\mathcal{B}_o)$, the memory-only ceiling.
\end{proposition}
\end{samepage}

The proposition holds because decoding is deterministic, records are encoded independently, and the workspace is the only input that crosses between modules (proof in Appendix~\ref{app:protocol-ext}). It bounds what the receiver's memory alone can support, and how far a trained model falls below this bound is measured in Section~\ref{sec:necessity}. Rung 2 then asks \emph{what} flows through the channel.}

\subsection{Rung 2: is the content used, or only relayed?}

\label{sec:counterfactual}

The second test keeps the channel and changes its content. We run the model on the recipient input and replace one packet of the source, its write-2 packet on the first token of the answer, with the packet the same module wrote on the donor input. Everything else evolves freely, including the competition for the workspace. The candidate computations now predict different outputs. Ignoring the transplant predicts room ($X$). Relaying the donor's answer predicts kitchen ($Y$), the outcome that models trained to relay produce (Section~\ref{sec:whocomputes}). Combining the transplanted content (Bill) with the receiver's own location memory predicts hospital ($Z$). In a standard two-value patch, $Y$ and $Z$ coincide, and an output that follows the donor cannot separate computing from relaying. \rev{Following \citet{variengien2025}, \citet{prakash2026}, and \citet{gurarieh2026},} we therefore evaluate only three-value cases, in which $X$, $Y$, and $Z$ are pairwise distinct. They are selected symbolically from the worlds and never from model outputs. Transplanted packets are the donor's own activations, so no correct answer is ever inserted. Controls replace the packet with itself and transplant at the never-read write 3 (Appendix~\ref{app:interventions}).

\rev{\textbf{Calibration: the test reports relay when relay is the computation.} We built two small reference networks, separate from AFGW, that fix where the computation happens and trained each with two seeds (Appendix~\ref{app:exp-reference}). In the relay design, the receiver's decoder sees only the source's packet, so the source must compute the answer. In the recombination design, the source sends the intermediate person and the receiver reads its own memory. All four models answer every held-out composition question, and all depend on their channel. The transplant separates them completely: the relay models give the donor's answer $Y$ and the recombination models the hybrid answer $Z$, in all 1,024 cases of every model (Figure~\ref{fig:teaser}, bottom).}

\subsection{Rungs 3 and 4: how does the content act, and for how many receivers?}

\label{sec:mechanism}

Content use leaves open how the content acts. Our hypothesis is that it changes what the receiver retrieves from its own memory: the broadcast redirects the location reader's query from Jeff's record to Bill's. The transplant enters the computation only through the broadcast $b^3$, which reaches the output either through the receiver's round-3 queries and the records they retrieve (route a) or directly through the receiver's final state (route b).

Two complementary interventions on the receiver's queries test route (a). \emph{Restoration} keeps the transplant but restores the receiver's round-3 queries to those of its native run. This closes route (a) and leaves route (b) open, so if the native output returns, redirected retrieval is necessary. \emph{Installation} (the query-only transplant) keeps the native broadcast but installs the round-3 queries that the transplant induced. If it reproduces the transplant's output, redirected retrieval is sufficient. With the native and transplant runs, these form a $2\times2$ design over broadcast and queries (Figure~\ref{fig:mechanism}a, Appendix~\ref{app:analysis-factorial}). Rung 4 applies rungs 2 and 3 to a variant in which one source packet reaches three receivers with different knowledge, as a global broadcast should.

\begin{table}[H]
\caption{\rev{What each rung changes and leaves alone, its prediction, and the alternative ruled out. Values after an intervention are recomputed.} ``Write 2'' and ``round 3'' are on the first answer token.}
\label{tab:rungs}
\centering\scriptsize
\setlength{\tabcolsep}{3pt}
\begin{tabularx}{\linewidth}{@{}>{\raggedright\arraybackslash}p{.1\linewidth}>{\raggedright\arraybackslash}p{.255\linewidth}>{\raggedright\arraybackslash}p{.15\linewidth}>{\raggedright\arraybackslash}X>{\raggedright\arraybackslash}p{.17\linewidth}@{}}
\toprule
Rung & Changes & Not intervened on & Prediction tested & Rules out\\
\midrule
1 Dependence & every received broadcast set to zero (OFF) & memories, reads, updates, output head & composition accuracy falls, local accuracy stays & an ignored channel\\
2 Content use & the source's write-2 packet replaced by its packet from the donor input & everything else (later steps evolve freely) & $X$ if the packet is ignored, $Y$ if relayed, $Z$ if the receiver recombines & relay of the donor's answer\\
3 Mechanism & the receiver's round-3 queries: native ones restored, or induced ones installed alone & broadcast, private state, memories, output head & restoration gives the native output, installation alone the transplant's output & an output change through the receiver's state alone\\
4 Breadth & one source packet read by three receivers (separate variant) & each receiver's own memory & each receiver gives its own $Z$, and rung 3 holds per receiver & use confined to one receiver\\
\bottomrule
\end{tabularx}
\end{table}

\section{Experimental Setup}

\label{sec:setup}

\textbf{Data and training.} A family fixes four objects, people and locations, and its 432 questions cross binding and location permutations, record orders, six question types (four local and two composition), and queried entities. Each model trains in stages, first on local questions and then on a mix of local and composition questions (Appendix~\ref{app:training}).

\rev{\textbf{Models.} We generated two corpora, A and B, each with 32 training families and 16 held-out evaluation families, and trained two initializations on each. We call these four the \emph{main models}. Their evaluation panels (a panel is a fixed evaluation set), intervention site (write 2 and read 3 on the first answer token) and metrics were fixed before the models were evaluated (Appendix~\ref{app:exp-main}). Four confirmation models, trained on 8 families per corpus under a plan fixed before training, give the same verdicts (Appendix~\ref{app:g4}). Table~\ref{tab:models} lists the systems we study.}

\begin{table}[H]
\caption{\rev{Systems in the main text, grouped by the question they answer. $^{\ast}$Trained on 8 families per corpus.}}
\label{tab:models}
\centering\scriptsize
\setlength{\tabcolsep}{4pt}
\begin{tabularx}{\linewidth}{@{}lr>{\raggedright\arraybackslash}Xll@{}}
\toprule
System & $n$ & Differs from the main models in & Figure & App.\\
\midrule
\multicolumn{5}{@{}l@{}}{\emph{How broadcast content is used (Sections~\ref{sec:necessity}--\ref{sec:reuse})}}\\
Main AFGW models & 4 & --- & \ref{fig:teaser}, \ref{fig:functional}, \ref{fig:mechanism}, \ref{fig:generality}b & \ref{app:exp-main}\\
No direct lexical read & 4 & the output head & \ref{fig:generality}b & \ref{app:exp-completeness}\\
Three receivers$^{\ast}$ & 2 & one source and three receivers & \ref{fig:functional}c, \ref{fig:generality}b & \ref{app:reuse}, \ref{app:exp-breadth}\\
\addlinespace[1.5pt]
\multicolumn{5}{@{}l@{}}{\emph{Who computes (Sections~\ref{sec:counterfactual} and \ref{sec:whocomputes})}}\\
Relay and recombination references$^{\ast}$ & 2+2 & small networks built to fix who computes & \ref{fig:teaser} & \ref{app:exp-reference}\\
One module reads both memories & 4+4 & memory access of the source or the receiver & --- & \ref{app:exp-whocomputes}\\
Relay-trained models and their control & 5+1 & memory access, output head, training data & \ref{fig:teaser} & \ref{app:exp-whocomputes}\\
Single-reader workspace & 2 & one module reads all records; a second task & \ref{fig:teaser} & \ref{app:exp-earlier}\\
\addlinespace[1.5pt]
\multicolumn{5}{@{}l@{}}{\emph{Whether the tests travel (Section~\ref{sec:beyond})}}\\
Dual-route$^{\ast}$ & 2 & a learned second route between the readers & \ref{fig:generality}a & \ref{app:exp-bypass}\\
Eight-entity worlds & 4 & evaluation only, no retraining & \ref{fig:generality}b & \ref{app:exp-scale}\\
Uniform routing & 4 & routing weights fixed at $1/5$ & \ref{fig:generality}b & \ref{app:exp-whocomputes}\\
Point-to-point, slot, order-invariant reader$^{\ast}$ & 2 each & the communication design or the reader & \ref{fig:generality}b & \ref{app:exp-otherarch}\\
Qwen2.5-1.5B-Instruct & 1 & a pretrained Transformer & \ref{fig:generality}b & \ref{app:exp-qwen}\\
\bottomrule
\end{tabularx}
\end{table}

\textbf{Panels and metrics.} The primary panel has 512 symbolically selected three-value cases per model. All metrics use the full panel, including cases the model answers wrongly without intervention, and require exact match of the complete answer and EOS. The hybrid rate is the fraction of outputs equal to $Z$. \rev{A second panel of new worlds repeats the tests (Appendix~\ref{app:exp-main}).}

\section{Results}

\label{sec:results}

\subsection{Rung 1 (dependence): the trained model depends on its workspace}

\label{sec:necessity}

\rev{The four main models answer 98\% of held-out composition questions (2,006 of 2,048) and 99\% of local questions. Switching the workspace off removes 69--78 points of composition accuracy and at most 2.3 points of local accuracy (Figure~\ref{fig:functional}a). The deficit is larger than the architecture requires. With the workspace off, the best predictor that uses the receiver's memory alone still answers 46--53\% of these questions (Proposition~\ref{prop:off}; band in Figure~\ref{fig:functional}a), and the models answer 21--27\%. How strongly a model comes to depend on its workspace is therefore learned (Appendix~\ref{app:protocol-ext}).}

\subsection{Rung 2 (content use): receivers compute with broadcast content}

\label{sec:content}

\begin{figure}[t]
\centering\includegraphics[width=\linewidth]{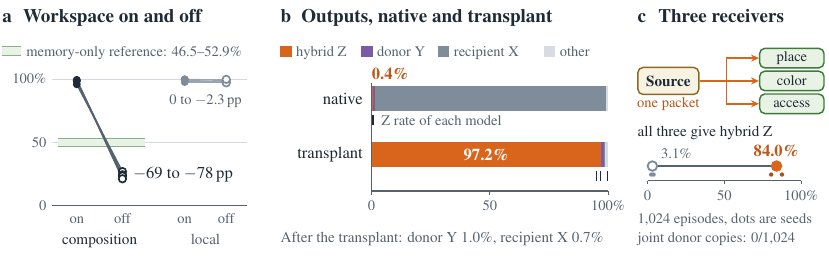}
\caption{\textbf{Communication supports receiver-specific computation.} (a) Accuracy with the workspace on and off for the four main models (512 questions per model and task). The band is the expected score of the best memory-only predictor for the two corpora. (b) Output categories on the 2,048 primary cases, natively and after the transplant. (c) Three receivers: rate at which all three give their hybrid answers jointly, natively and after one write-2 packet (1,024 episodes).}
\label{fig:functional}
\end{figure}

\rev{One transplanted packet yields the hybrid answer $Z$ in 97.2\% of cases (1,990 of 2,048), against 0.4\% without the transplant. The donor's answer $Y$ accounts for 1.0\% of outputs and the original answer $X$ for 0.7\% (Figure~\ref{fig:functional}b), so the receiver computes with the transplanted content. Scoring relay against the donor's actual output gives the same verdict (Appendix~\ref{app:exp-main}).}

\subsection{Rung 3 (mechanism): content acts by redirecting the receiver's retrieval}

\label{sec:retrieval}

\begin{figure}[t]
\centering\includegraphics[width=\linewidth]{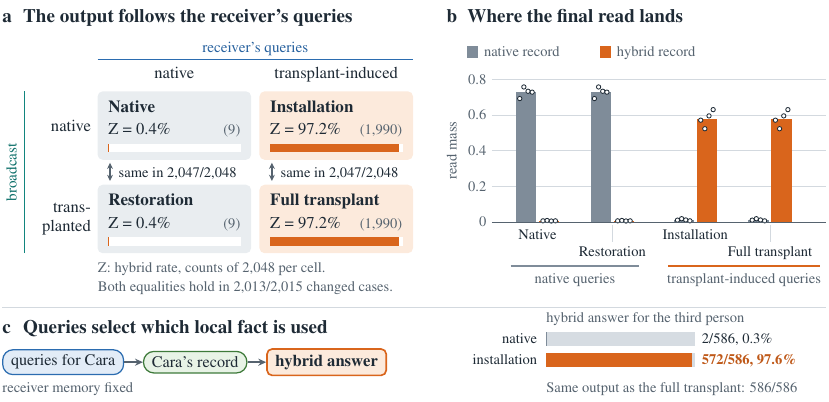}
\caption{\textbf{The receiver's queries recover and reproduce the transplant's effect.} (a) Hybrid rates in the $2\times2$ design over broadcast and queries (2,048 cases per cell). Arrows give the agreement of complete outputs. (b) Final-read mass on the native and hybrid records in four conditions. (c) With the receiver's memory fixed, queries induced by a second donor yield the corresponding hybrid answer.}
\label{fig:mechanism}
\end{figure}

The broadcast acts as a key: it tells the receiver whose record to retrieve (Figure~\ref{fig:mechanism}a). \textbf{Redirected retrieval is necessary.} With the transplanted packet still in place, restoring only the receiver's round-3 queries recovers the complete native output in 2,047 of 2,048 cases, while the direct route into the receiver's state stays open. \textbf{Redirected retrieval is sufficient.} With the native broadcast in place, installing only the transplant-induced queries reproduces the transplant's complete output in \rev{2,047} of 2,048 cases. \rev{Both read tests hold in 2,013 of the 2,015 cases the transplant changes. The effect is specific to the final read: random query changes of the same size leave 93--97\% of it in place, and with all three source packets transplanted, restoring the round-2 queries instead undoes none of the changes. The same holds on a second panel and on chains never seen in training (Appendix~\ref{app:exp-main}).}

\textbf{The queries name the person to look up.} Under the transplant, the receiver's final read shifts from Jeff's record to Bill's (read mass, the weight a read places on a record, rises on the record needed for $Z$ from \rev{0.01 to 0.52--0.63}; Figure~\ref{fig:mechanism}b), and restoration shifts it back. \rev{When a second donor gives the football to a third person, Cara, installing only the queries its packet induces yields Cara's hybrid answer in 572 of 586 cases, against 2 natively (Figure~\ref{fig:mechanism}c).}

\rev{\textbf{Retrieval persists without the direct lexical read.} AFGW's output head reads the receiver's final lexical read directly, which could favor retrieval. Four matched models without this direct read compose equally well, and the read tests over reads 1--3 hold in 98.8--100\% of the outputs that a three-packet transplant changes (Figure~\ref{fig:generality}b, Appendix~\ref{app:exp-completeness}).}

\subsection{Rung 4 (breadth): one broadcast packet serves three receivers}

\label{sec:reuse}

\rev{We trained a variant (two seeds) in which one source that knows who holds what feeds three receivers that know people's locations, colors and access rights, so one source packet implies a different hybrid answer at each receiver (Appendix~\ref{app:reuse}). Its workspace is computed once per input and read by all three. On a panel from four new families, the source's write-2 packet alone makes all three receivers give their own hybrid answers jointly in 84\% of episodes, against 3\% natively, and no episode copies the donor's joint output (Figure~\ref{fig:functional}c). Rung 3 holds receiver by receiver: both read tests hold in 2,894 of the 2,926 outputs the packet changes, and intervening on one receiver leaves the other two unchanged (Appendix~\ref{app:exp-breadth}).}

\subsection{Who computes is learned: recombination, relay, or a nearly unused broadcast}

\label{sec:whocomputes}

\rev{\textbf{Training selects recombination where relay is available.} In eight further models one module reads both memories and could compute the answer alone (the source in four, the receiver in four). All eight still give $Z$ in 97\% of cases, and when records are replaced at inference the answer follows the receiver's memory (98\%), not the source's copy (under 1\%).}

\rev{\textbf{The same architecture learns relay.} Two models, in which the source reads both memories and the output head has no direct lexical read, differ only in their training data. In one, the receiver's records, which hold the answer, are resampled at random during training, so only the source can compute it. Both answer at least 511 of 512 held-out questions and fail without the workspace. A standard patch, whose donor answer is also the hybrid answer, moves both to the donor's answer in every case. The three-value transplant (writes 0--2) gives the donor's answer in 511 of 512 cases in the first and the hybrid answer in all 512 in the second. Replacing either module's records at inference confirms both verdicts. Four further models, two per corpus, whose receiver's records are resampled on only a quarter of the training examples, relay in 428 to 512 of 512 cases (Appendix~\ref{app:exp-whocomputes}).}

\rev{\textbf{A built-in broadcast can go nearly unused.} We trained a workspace model like AFGW, with four recurrent modules, four rounds and a competitive broadcast, but in which one module reads all records and the output head also reads the workspace. It learns a second two-hop task, posed as short narratives (Appendix~\ref{app:exp-earlier}). Two such models answer every held-out question, half of them compositions. The protocol shows that their modules do so without the broadcast: zeroing everything the modules receive from it leaves 2,017 of 2,048 answers correct, and a donor's broadcast in all four rounds changes none of 2,048 outputs. The one reader chains its own reads, so nothing crosses between modules. AFGW's two boundaries, split memories and no path from the workspace to the output, close this bypass, and what then crosses the workspace is still learned.}

\subsection{Beyond an exclusive channel: replaceable routes and other interfaces}

\label{sec:beyond}

\begin{figure}[t]
\centering\includegraphics[width=\linewidth]{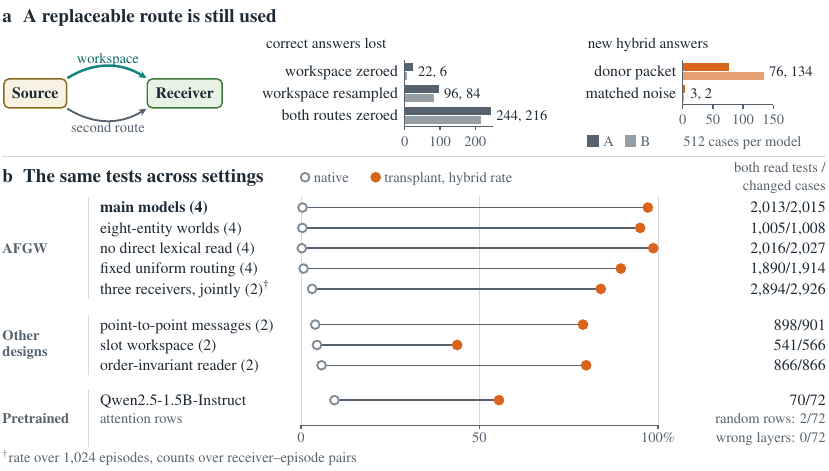}
\caption{\rev{\textbf{The content and read tests across routes, designs and interfaces.} (a) The two dual-route models, one per corpus: correct answers lost when the workspace is zeroed, resampled from another input, or zeroed together with the second route, and new hybrid answers from the donor packet and from matched noise. (b) Hybrid rates natively and after the transplant, with the main models as reference. Counts on the right give the changed cases in which both read tests hold. The row without the direct lexical read uses writes 0--2 and reads 1--3. Table~\ref{tab:models} lists the models.}}
\label{fig:generality}
\end{figure}

\rev{\textbf{A replaceable route still contributes to computation.} We added a learned second route between the two memory readers and trained this dual-route model with the workspace zeroed on half of its training examples, so that either route can sustain performance (two models; Appendix~\ref{app:exp-bypass}). Zeroing the workspace costs only 22 and 6 of 512 answers, so zero-ablation would call it nearly unnecessary. Yet its transplant moves 76 and 134 outputs to the hybrid answer, against 3 and 2 for matched packet noise, and the read tests hold in 305 of 309 changed cases.}

\rev{\textbf{The signature holds across scale and designs.} By the \emph{signature} we mean a transplant that raises hybrid answers together with both read tests (Figure~\ref{fig:generality}b). The main models, never retrained, carry the mechanism to eight-entity worlds, where the transplant raises hybrid answers from 0.4\% to 95\% (Appendix~\ref{app:exp-scale}). The tests also transfer unchanged to other designs, among them point-to-point messages between modules and a slot-based workspace in the style of \citet{goyal2022} (Appendix~\ref{app:exp-otherarch}). In every setting, both read tests hold in at least 95\% of changed cases.}

\rev{\textbf{A pretrained Transformer.} We ran rungs 2 and 3 on Qwen2.5-1.5B-Instruct \citep{qwen2024} at a known interface, a name that the model generates and a later turn looks up in a table of codes. Transplanting the name's representation raises hybrid answers from 12 to 71 of 128. The read tests act on the attention rows of the answer position, in the layers where it attends to the table, and jointly succeed in 70 of 72 changed cases, against 2 of 72 for matched random rows (Appendix~\ref{app:exp-qwen}).}

\section{Related Work}

\label{sec:related}

\textbf{Consciousness science and AI.} GWT is among the theories from which indicator properties for AI are derived \citep{butlin2023}, and philosophical analyses ask whether language models or agents meet its conditions \citep{chalmers2023,goldstein2024}. Such criteria establish that a workspace is present. We contribute a test of whether its content is used and a model organism in which the test is exact, as argued for toy models of consciousness theories \citep{albantakis2025}.

\textbf{Workspaces in AI.} Trained neural workspaces aid coordination, generalization, and transfer \citep{goyal2021,goyal2022,vanrullen2021,devillers2024,maytie2024,dossa2024,chateau2025}, but they are evaluated by performance and ablation, which show that a workspace helps and leave open how it is used. \rev{Learned channels that are necessary yet not used as intended are well documented, from answer codes in emergent communication \citep{kottur2017,lowe2019} to labels leaked through rationales and concept bottlenecks \citep{yu2019,havasi2022}.} \citet{gurnee2026} discover workspace-like representations in pretrained LLMs and study their downstream roles, an analytic route that retains scale, whereas our constructive route makes every pathway enumerable (Appendix~\ref{app:related}).

\textbf{Causal analysis of neural networks.} The protocol builds on activation patching and mediation \citep{vig2020,meng2022}, interchange interventions and causal abstraction \citep{geiger2021,geiger2024}, and path patching \citep{wang2023,goldowskydill2023}. \rev{\citet{zhang2024} show that patching choices can change conclusions. As they advise, we use real-input donors and vary ablation and metric, with the same verdicts (Appendix~\ref{app:exp-off}). Distinct-output counterfactuals appear in causal-abstraction work on retrieval and binding \citep{variengien2025,prakash2026,gurarieh2026,mueller2025}. We turn this design into a symbolic inclusion rule, fixed before any output is seen, calibrate it on systems whose computation is known, and pair it with interventions on the receiver's reads. The combination answers a question that a two-value patch leaves open: who computes with communicated content, and how (Section~\ref{sec:whocomputes}).}

\section{Discussion and Conclusion}

\label{sec:discussion}

\rev{\textbf{What the architecture fixes and what training selects.} The architecture fixes which information each module can see, not how it is used. The size of the OFF deficit and the route through retrieval are both learned (Sections~\ref{sec:necessity} and~\ref{sec:retrieval}), and so is the division of work: the same modules learn to recombine or to relay depending on training data, and a single-reader workspace leaves its broadcast nearly unused (Section~\ref{sec:whocomputes}). Use is learned even where it is optional: a receiver that reads both memories and could compose alone still loses 75 points under OFF (Appendix~\ref{app:exp-whocomputes}).}

\rev{\textbf{The content and read tests travel beyond an exclusive pathway.} Table~\ref{tab:requirements} (Appendix~\ref{app:discussion-ext}) states what each rung requires. Only the bound of Proposition~\ref{prop:off} needs the workspace to be the only cross-module input. The content and read tests need a carrier and a reader, not an exclusive wire, and they detect use even where a second route compensates and ablation shows almost no deficit (Section~\ref{sec:beyond}).}

\textbf{Implications for consciousness science.} GWT's central claim is functional: broadcast content is consciously accessible because consumers use it. Indicator-based assessments of AI systems ask whether a system \emph{has} a workspace \citep{butlin2023}. Our results show that this underdetermines the computation\rev{, so whether broadcast content is used, and how, has to be tested.} We suggest that claims of workspace-like organization, designed or discovered, be supported by content-level and route-specific causal evidence. \rev{Use does not depend on how access is allocated: it holds with learned and with uniform routing (Table~\ref{tab:commitments}).} Further GWT properties such as ignition, capacity limits, maintenance and report can be tested with the same logic, and workspace-like representations discovered in large models \citep{gurnee2026} are the natural next target.

\textbf{Conclusion.} Building a theory-mandated component is half the task. The other half is showing what the learned system computes with it. In a model organism built to make these distinctions exact, broadcast content is used and acts by redirecting the receiver's retrieval, and the same signature appears across scales, routes, designs and a pretrained Transformer. We offer the protocol, with AFGW as its reference instantiation, as a reusable standard for turning the functional claims of consciousness science into falsifiable properties of trained systems.

\label{maintextend}
\clearpage
\setlength{\textfloatsep}{\vthirtyninestextfloatsep}
\setlength{\floatsep}{\vthirtyninesfloatsep}
\setlength{\intextsep}{\vthirtyninesintextsep}
\subsection*{AI Use Statement}

Generative AI tools were used for grammatical correction and language polishing of author-written text to improve clarity and readability.

\section*{Reproducibility Statement}
Appendix~\ref{app:architecture} specifies the architecture, Appendix~\ref{app:training} the data and training, Appendix~\ref{app:exp-main} the main models, and Appendix~\ref{app:interventions} the intervention definitions and metrics. Table~\ref{tab:models} lists the systems of the main text and the appendix that reports each.

\bibliographystyle{plainnat}
\bibliography{references}
\clearpage
\appendix
\setlength{\textfloatsep}{12pt plus 2pt minus 2pt}
\setlength{\floatsep}{10pt plus 2pt minus 2pt}
\setlength{\intextsep}{6pt plus 2pt minus 2pt}
\renewcommand{\floatpagefraction}{0.85}
\section{Notation and Reader's Guide}

\label{app:guide}

This guide says where each piece of evidence is reported and collects the notation.

\rev{\textbf{Models.} The main text reports the four main models (A3, A4, B3 and B4) and the variants listed in Table~\ref{tab:models}. The main models are trained on 32 families per corpus for 12,288 updates (Appendix~\ref{app:exp-main}). Four further models with the same architecture, corpora and evaluation panels, trained on 8 families per corpus for 4,096 updates under a plan fixed before training, are the confirmation models (A1, A2, B1 and B2). They give the same verdicts, and Appendix~\ref{app:g4} collects them together with every analysis run on them.}

\textbf{Where to find the evidence.} The appendix has two parts. Appendices~\ref{app:extended}--\ref{app:training} and \ref{app:interventions}--\ref{app:analysis} support the main text: extended related work and discussion (\ref{app:extended}), architecture (\ref{app:architecture}), data and training (\ref{app:training}), intervention definitions and the proof of Proposition~\ref{prop:off} (\ref{app:interventions}), the three-receiver variant (\ref{app:reuse}), and analytical properties (\ref{app:analysis}). Appendix~\ref{app:exp} reports the experiments on the main models, their variants and other systems in the order of Section~\ref{sec:results}, and Appendix~\ref{app:g4} reports the confirmation models. Table~\ref{tab:exp-status} lists the question each experiment answers and where it is reported.

\begin{table}[!htbp]
\caption{Questions answered by the experiments of Appendices~\ref{app:exp} (top) and~\ref{app:g4} (bottom), and where each is reported.}
\label{tab:exp-status}
\centering\small
\begin{tabular}{@{}L{.84\linewidth}l@{}}
\toprule
Question & Section\\
\midrule
Do rungs 1--3 hold in models that answer 98\% of composition questions? & \ref{app:exp-main}\\
Does retrieval persist without the direct lexical read? & \ref{app:exp-completeness}\\
Rung 4 with a single packet, different-map donors, new families, and rung 3 per receiver & \ref{app:exp-breadth}\\
Does training ever select relay, and does the protocol detect it? Do matched variants at high accuracy recombine? & \ref{app:exp-whocomputes}\\
Does a trained workspace model ever leave its broadcast nearly unused? What are AFGW's boundaries for? & \ref{app:exp-earlier}\\
Does the three-value test report relay when relay is the mechanism? & \ref{app:exp-reference}\\
What does each rung license without an exclusive channel? & \ref{app:exp-bypass}\\
Do rungs 1--3 carry over to worlds with twice as many entities? & \ref{app:exp-scale}\\
Do the rungs run on other communication architectures and an order-invariant reader? & \ref{app:exp-otherarch}\\
Can rungs 2 and 3 be run in a pretrained Transformer? & \ref{app:exp-qwen}\\
\midrule
Is rung 1 robust to the ablation method, and how does it compare with the memory-only reference? & \ref{app:exp-off}\\
Does rung 2 exclude relay on actual donor outputs, where a two-value patch cannot? & \ref{app:exp-relay}\\
Do rungs 2 and 3 replicate on unseen panels? Does the redirected query carry a specific entity? & \ref{app:exp-fresh}\\
What do the routing weights $\pi$ do at the transplant site? Does a model trained with uniform routing keep the signature? & \ref{app:exp-routing}\\
When relay is available, does training select it, and does the three-value verdict match lesion ground truth? & \ref{app:exp-relayarms}\\
\bottomrule
\end{tabular}
\end{table}

\textbf{Worlds and answers.} The recipient world $(B_0,S_1)$, with bindings $B$ and locations $S$, is the input on which the model is run and intervened on. The donor world $(B_1,S_0)$ is the input from which the transplanted packet is taken. The hybrid world $(B_1,S_1)$ combines the donor's bindings with the recipient's locations. This notation is written for object$\to$location questions, where the source holds the bindings. For location$\to$object questions the two relations exchange roles. $X$, $Y$, and $Z$ are the correct answers in the recipient, donor, and hybrid worlds. The primary panel contains only cases in which $X$, $Y$, and $Z$ are pairwise distinct.

\textbf{Modules and sites.} The source is the binding reader $E_1$ for object$\to$location questions and the location reader $E_2$ for location$\to$object questions, and the receiver $o$ is the other reader. Write $r$ is the packet written in round $r$, which enters the broadcast $b^{r+1}$. Read $r$ is the receiver's memory read in round $r$, whose queries depend on $b^r$. The primary site is write 2 followed by read 3, both on the first answer token.

\textbf{Interventions.} The appendix uses the names of the main text. OFF (Definition~\ref{def:off}) zeroes every module's received broadcast in every round of every generated token. The transplant replaces the source's packet. Restoration keeps the transplant and restores the receiver's native queries. Installation, also called the query-only transplant, keeps the native broadcast and installs the queries that the transplant induced. Query noise keeps the transplant and changes the receiver's queries by a random displacement with the norm of the restoration displacement. Appendix~\ref{app:analysis-factorial} indexes the four runs of the $2\times2$ design by broadcast and queries: native 00, transplant 11, restoration 10, and installation 01.

\textbf{Metrics.} The hybrid rate is exact match to $Z$ followed by EOS. Hybrid agreement is identity with the same model's native output on the hybrid world, errors included. Native recovery is identity with the unperturbed output.

\textbf{Routes (Figure~\ref{fig:intervention}, right).} Route (a) runs from $b^3$ through the receiver's round-3 queries and reads to the output. Route (b) runs from $b^3$ directly into the receiver's final private state.

\section{Extended related work and discussion}
\label{app:extended}

\subsection{Extended related work}
\label{app:related}
\textbf{Consciousness science meets AI.} GWT and its neuronal variant \citep{baars1988,dehaene1998,dehaene2011,mashour2020} are among several theories now used to derive functional criteria for assessing AI systems \citep{butlin2023}. Following \citeauthor{block1995}'s (\citeyear{block1995}) distinction, these criteria concern access consciousness, that is, the availability of information for reasoning, report, and control. Philosophical analyses have asked whether LLMs or LLM-based agents meet GWT's conditions \citep{chalmers2023,goldstein2024}. The consciousness prior \citep{bengio2017} and \citet{juliani2022} link conscious-access functions to learning and generalization, and \citet{kanai2026} recently proposed a control-theoretic criterion that identifies a workspace as a mediating subnetwork. Our contribution complements these analyses. It specifies what evidence shows that a learned system uses a theory-mandated organization, and it supplies a model organism in which that evidence is decisive. This follows the broader case for toy models as a way to make theories of consciousness concrete and testable \citep{albantakis2025}.

\textbf{Scaffolded workspaces.} Classical cognitive architectures implemented GWT-like cycles symbolically \citep{shanahan2006,franklin2014}, and their modern successors orchestrate pretrained LLMs around a central buffer. The Unified Mind Model builds an autonomous agent around GWT \citep{hu2025}, Global Workspace Agents run an event-driven broadcast among heterogeneous LLM agents \citep{shang2026}, and \citet{goldstein2024} argue that such language agents may satisfy GWT's conditions. In all of these systems, the workspace is external text passed between model calls, so the communication interface is specified by design.

\textbf{Trained neural workspaces.} \citet{goyal2022} replaced pairwise communication among neural modules with a limited-capacity shared workspace, reporting gains across tasks that include a next-word prediction experiment. Related modular designs include Recurrent Independent Mechanisms \citep{goyal2021}. The Global Latent Workspace \citep{vanrullen2021} has been realized as a shared multimodal latent space that supports semi-supervised learning and zero-shot policy transfer \citep{devillers2024,maytie2024}. \citet{dossa2024} embodied a workspace agent in a multimodal environment, and \citet{chateau2025} routed information through a workspace to chain operations. These works establish that workspace bottlenecks can be useful. AFGW differs from them in three respects. It is an autoregressive language generator. Its information boundaries make the broadcast the only possible cross-module route, so \rev{what switching it off can achieve is bounded by a provable receiver-memory ceiling (Proposition~\ref{prop:off}, Corollary~\ref{cor:ceiling}).} And it comes with causal tests that ask what broadcast content does inside the receiver. \rev{Section~\ref{sec:beyond} applies the same tests to a slot-based workspace in the style of \citet{goyal2022}.}

\textbf{Discovered workspaces.} \citet{gurnee2026} use a Jacobian lens to identify a small set of verbalizable representations in pretrained LLMs that show functional hallmarks of a global workspace, including use by downstream computations, a striking finding because these models were never designed to have one. Their analytic route and our constructive route are complementary. The constructive route brings the architecture closer to the one GWT describes, with specialized modules, competition for access, and recurrent broadcast as explicit components. The workspace is a specified part of the computation, so its identity does not depend on an analysis tool. And because the broadcast is the only route between modules, interventions on it change exactly the information we intend to change, whereas a candidate workspace that shares the residual stream with all other computation may leave redundant routes open. The discovered-workspace approach offers scale and naturalness, and the constructive approach offers exact control. The content and read tests apply to a discovered representation once its consumers are identified.

\rev{\textbf{Channels that are not used as intended.} Trained global-workspace models are evaluated by performance, ablation and transfer, which establish task success and dependence on the channel, not what the channel carries. Whether such a model uses its broadcast as intended has therefore gone untested, and Appendix~\ref{app:exp-earlier} reports a trained workspace model that leaves its broadcast nearly unused. Elsewhere in machine learning, learned channels that are necessary for the task yet carry something other than what they were designed to carry are well documented. Agents trained to communicate develop answer codes instead of compositional messages \citep{kottur2017}, and standard measures can credit a speaker with informative messages that barely change what the listener does \citep{lowe2019}. Fine-tuned translation agents drift away from the pivot language they were meant to use \citep{lee2019}. Rationales can leak the label instead of explaining it \citep{yu2019}, concept bottlenecks can pass task information through soft concept scores \citep{havasi2022}, and the modules of compositional neural module networks can stop performing their intended sub-functions while the whole still answers correctly \citep{subramanian2020}. In each case an ablation would show that the channel is needed, and only a test of content would show that it is not used as intended. AFGW fixes which module reads which memory, but what the broadcast carries is learned, so the same gap applies.}

\textbf{Causal analysis of neural computation.} Our interventions build on activation patching and causal mediation analysis \citep{vig2020,meng2022,zhang2024}, interchange interventions and causal abstraction \citep{geiger2021,geiger2024}, path-level patching \citep{wang2023,goldowskydill2023}, and the distinction between direct and indirect effects \citep{pearl2001}. Recent work cautions that interpretability claims can outrun their evidence: candidate explanations may be non-identifiable \citep{meloux2025}, plausible-looking findings can arise even in randomly initialized networks \citep{meloux2025salmon}, interventions can push representations off the model's natural distribution \citep{grant2026}, and counterfactual claims are hard to verify without controlled supervision \citep{joshi2026}. Our protocol is designed with these concerns in mind. Every rung is tested against an explicit alternative hypothesis, transplanted packets are the donor's native activations, symbolic worlds supply exact counterfactual targets, and panels, sites and metrics are fixed before evaluation. \rev{Choosing counterfactuals whose competing hypotheses predict distinct outputs is established practice in causal-abstraction work: \citet{variengien2025} track the recipient's, the donor's, and a hybrid label; \citet{prakash2026} use patches whose predicted answer matches neither run; \citet{gurarieh2026} score pairwise-distinct predictions with a separate bin for mixed outputs; and MIB \citep{mueller2025} shows how a correlated output variable can masquerade as an intermediate one. We adopt this design. Our addition is to make pairwise-distinct $X$, $Y$, $Z$ a symbolic inclusion rule, fixed before any output is seen, to score relay against the donor's actual output as well as its symbolic answer, and to calibrate the verdict on systems whose computation is known. \citet{zhang2024} study a complementary question, how the corruption method, metric, and patching window affect patching conclusions.} Because the architecture enforces its information boundaries, we can also enumerate all paths from the intervention point to the output (Section~\ref{sec:mechanism}), which makes query restoration a sharp test of which route carries the effect.

\textbf{Compositional reasoning.} Our tasks are two-hop relational compositions. In this setting, neural sequence models struggle with systematic generalization \citep{lake2018}, and LLMs show a gap between knowing facts and composing them \citep{press2023,yang2024,wang2024}. We use composition as a lever, not as a benchmark: it forces information held by one module to be used by another.

\Needspace{22\baselineskip}
\subsection{Extended discussion}
\label{app:discussion-ext}
\begin{table}[!htbp]
\caption{\rev{What each rung requires and supports, with and without an exclusive route.}}
\label{tab:requirements}
\centering\scriptsize
\setlength{\tabcolsep}{3pt}
\begin{tabularx}{\linewidth}{@{}>{\raggedright\arraybackslash}p{.12\linewidth}>{\raggedright\arraybackslash}X>{\raggedright\arraybackslash}X>{\raggedright\arraybackslash}X@{}}
\toprule
Rung & Requires & Supports & Without an exclusive route\\
\midrule
1 Dependence & a channel that can be switched off, and exclusivity for the ceiling of Prop.~\ref{prop:off} & dependence on the channel & a deficit still shows dependence, but a small deficit no longer implies little use\\
2 Content use & an identifiable carrier, and exact targets with $X$, $Y$, $Z$ distinct & use of the carried content by the receiver, not relay & informative for the content that carrier delivers\\
3 Mechanism & an identifiable read interface on the receiver & redirected reads are necessary and sufficient for the change & locates where content is consumed\\
4 Breadth & the requirements of rungs 2 and 3 at each receiver & one packet serves several receivers & as for rungs 2 and 3\\
\bottomrule
\end{tabularx}
\end{table}

\textbf{Beyond GWT.} GWT is our test case, and the verification gap arises whenever a theory-mandated component is built into a learned system. The other theories surveyed by \citet{butlin2023} also yield indicators stated as functional roles of components, and implementing any of them opens the same distance between presence and use. Within GWT, we test one functional property at the heart of conscious access: broadcast content is globally available for other processes to use. We ask whether that property can be built into a language model, acquired through learning, and verified causally.

\textbf{Interpretability by design.} An explicit broadcast with enforced information boundaries offers interpretability by design. For any computation, one can state which module could have known what and test which route carried the answer. Model organisms with exact counterfactual ground truth can in turn serve as testbeds for interpretability methods, whose conclusions are otherwise hard to validate \citep{meloux2025salmon,joshi2026}.

\textbf{Toward scale.} \rev{Rungs 2--4 are stated in terms of an identifiable carrier, exact counterfactual targets, and a receiver-side read interface, not in terms of AFGW's modules, so they apply to other architectures that provide these (Table~\ref{tab:requirements}). We have validated the protocol in AFGW and related variants at 6.9M parameters and at a known interface of a pretrained 1.5B-parameter Transformer (Appendix~\ref{app:exp-qwen}).} A natural next step is to insert an AFGW-style workspace between specialist or retrieval components of a pretrained Transformer, where exact three-value counterfactuals remain constructible with synthetic knowledge bases.

\section{Architecture specification}
\label{app:architecture}
\subsection{From theory to design: commitments and overview}
\label{app:design}
We use GWT to choose architectural commitments, and we choose their implementation so that each rung of the protocol in Section~\ref{sec:protocol} can be applied exactly (Figure~\ref{fig:architecture}). We also explicitly separate what we supply by design from what training must discover (Table~\ref{tab:commitments}). We supply each module's memory access and specify which module produces the answer for each question type. Everything else must be learned from final-answer supervision alone: what to communicate, when to communicate it, and how to use what is received. Intermediate entities, supporting records, and routing sequences are never supervised.

\begin{table}[!htbp]
\caption{GWT commitments, their implementation, and what our experiments test.}
\label{tab:commitments}
\centering\small
\begin{tabular}{@{}L{.20\linewidth}L{.39\linewidth}L{.35\linewidth}@{}}
\toprule
GWT function & Implementation in AFGW & What is supplied vs. established\\
\midrule
Specialized, parallel processing & Four recurrent modules with private states. $E_1$ reads only binding records; $E_2$ reads only location records; $E_0$ and $E_3$ have no memory access. & Supplied: memory types and the answering module. Learned: the local tasks.\\
Limited-capacity shared access & One 640-dimensional packet per round, allocated by learned competition that includes a NULL option. & Established: the channel is used (Section~\ref{sec:necessity}). \rev{Use holds with learned and with uniform routing: a matched model trained with fixed uniform routing retains rungs 1--3 (Appendices~\ref{app:exp-whocomputes} and~\ref{app:exp-routing}).}\\
Global broadcast & The same message reaches every module before its next read and update. & Established: content-specific effects (Section~\ref{sec:content}); one broadcast serving three receivers (Section~\ref{sec:reuse}).\\
Content-dependent use by receivers & The received broadcast and the private state jointly form each receiver's memory queries. & Established: restoring the receiver's queries abolishes the content effect in all four main models, and installing the induced queries alone reproduces it (Section~\ref{sec:retrieval}).\\
\bottomrule
\end{tabular}
\end{table}

\textbf{Local encoding with enforced boundaries.} The base model, AFGW as used in the main text, has 6,891,784 parameters, an 8,192-token vocabulary, and 256-dimensional token representations. Two causal-attention layers with ALiBi biases \citep{press2022} encode each fact record independently, while the question and the generated prefix form a separate segment. We denote the current representation of that segment by $u_t$, and we denote a fixed projection of the question anchor by $g$. Because each record is encoded in isolation, no information crosses between memory banks inside the encoder. Four recurrent modules receive the public features $u_t$ and $g$, and the public question type determines the answering module, the receiver $o \in \{1,2\}$.

\Needspace{19\baselineskip}
\textbf{One internal round.} For each predicted token, the private states $h_e$ and the workspace $b$ start at zero, and four internal rounds $r = 0,1,2,3$ follow. In each round, a fact-reading module $e$ first forms two 128-dimensional queries and reads from its own memory bank $\mathcal{B}_e$ (Eq.~\ref{eq:query}). The contextual read $C$ and the lexical read $\ell$ use the same joint record/token weights; their values come from contextual states and from normalized token embeddings, respectively. Record selection is sigmoid-gated and order-sensitive: it favors the last matching record, and any unallocated mass becomes an explicit NULL read (Section~\ref{app:coupled} gives the exact reader). After a GRU-style update of its private state \citep{cho2014}, each module writes a packet consisting of a 128-dimensional control block and two 256-dimensional lexical blocks, the latter being zero for non-readers. The packets then compete for the workspace (Eq.~\ref{eq:broadcast}). Here the routing distribution $\pi^r$ is a softmax over the four modules and NULL, whose packet is zero. Its query depends on the goal and the previous workspace control state, while the module keys depend on the updated private states. Crucially, every module receives the same $b^{r+1}$, and this broadcast enters every reader's next query. The write at $r=3$ is not used in any later step, so it serves as a built-in negative control.

\textbf{Generation without a workspace shortcut.} A tied-embedding vocabulary head receives only four inputs: $g$, the receiver's final state $h_o^4$, the prefix representation $u_t$, and the receiver's final lexical reads. It has no direct input from the workspace $b$, from any other module's state, or from an unrestricted read of the full memory. Decoding is greedy over the full vocabulary and must terminate with an EOS token, without any candidate-answer restriction or output repair. Record encodings may be cached, but private and workspace states reset at every token, so the four-round core reruns for each generated token.
\subsection{Base-model dimensions and answering modules}
Table~\ref{tab:dimensions} specifies the base model used for the main capability and query-path experiments. A four-entity world has eight records and an eight-entity world sixteen. The three-receiver variant changes the fact interfaces and the decoding schedule (Appendix~\ref{app:reuse}).

\begin{table}[!htbp]
\caption{Base architecture.}
\label{tab:dimensions}
\centering\small
\begin{tabular}{@{}lr@{\qquad}lr@{}}
\toprule
Quantity & Value & Quantity & Value\\
\midrule
Trainable parameters & 6,891,784 & Vocabulary & 8,192\\
Token/model dimension & 256 & Private-state dimension & 192\\
Control dimension & 128 & Retrieval key dimension & 128\\
Modules & 4 & Internal rounds per token & 4\\
Reader heads per reader & 2 & Packet dimension & 640\\
Local encoder layers & 2 & Attention heads per layer & 4\\
Feed-forward dimension & 512 & Local attention window & 32\\
Maximum records & 16 & Tokens per record & 32\\
Maximum question tokens & 64 & Generation limit & 32\\
Logit scale & $1/16$ & Training arithmetic & FP32\\
\bottomrule
\end{tabular}
\end{table}

The public grammar assigns the four local questions and two composition questions to their receivers. Person-to-location and location-to-person use $E_2$; person-to-object and object-to-person use $E_1$. Object-to-location uses $E_2$, while location-to-object uses $E_1$.

\FloatBarrier
\subsection{Independent local encoding and coupled retrieval}
\label{app:coupled}
Each record is flattened into an independent batch element before the local encoder. ALiBi positions restart within each segment. Contextual token states are $H_{av}\in\R^{256}$, for record $a$ and token $v$. Learned pooling forms a record representation; learned maps form record keys $k_a$, token keys $K_{av}$, and contextual values $V_{av}$. A lexical value is instead a normalized raw embedding:
\begin{equation}
L_{av}=\sqrt{256}\,\frac{E(x_{av})}{\max(\|E(x_{av})\|_2,\epsilon)}.
\end{equation}
The question/prefix is a separately encoded segment. Its last valid position provides $u_t$, while the fixed question anchor $u_q$ gives $g=\tanh(W_g u_q)$.

Let $V_e(a)\in\{0,1\}$ be the visibility mask. Each reader uses two queries and the following ordered record selector:
\begin{align}
m^r_{eha}&=V_e(a)\,\sigma\left((q^r_{eh})^\top k_a/\sqrt{128}+\beta_{eh}\right),\\
w^r_{eha}&=m^r_{eha}\prod_{j>a}(1-m^r_{ehj}),\qquad
w^r_{eh,\NULL}=\prod_a(1-m^r_{eha}).
\label{eq:recordweights}
\end{align}
The last matching record is favored; the selector is not permutation invariant. A softmax over valid tokens within a visible record gives
\begin{equation}
\alpha^r_{ehav}=\operatorname{softmax}_{v\in\operatorname{valid}(a)}
\big((q^r_{eh})^\top K_{av}/\sqrt{128}\big).
\end{equation}
The same joint weights produce both outputs:
\begin{equation}
C^r_{eh}=\sum_{a,v}w^r_{eha}\alpha^r_{ehav}V_{av},\qquad
\ell^r_{eh}=\sum_{a,v}w^r_{eha}\alpha^r_{ehav}L_{av}.
\end{equation}
Padding and invisible records have zero weight, and the record weights sum to $1-w_{\NULL}$.

\subsection{Private recurrence, writes, and the vocabulary head}
The implemented query dependence can be written as
\begin{align}
f_e^r&=W_e^f[h_e^r;g;b^r],\qquad x_e^r=W_e^u[u_t;g],\\
q_e^r&=\operatorname{reshape}_{2\times128}W_e^q[h_e^r;f_e^r;u_t],\\
h_e^{r+1}&=\operatorname{GRU}_e\big(\tanh(x_e^r+f_e^r+\rho_e^r),h_e^r\big),
\end{align}
where $\rho_e^r$ is a projection of the two contextual reads for fact readers and zero for non-readers. Maps are module-specific, implemented in grouped form. For a reader, the control block $c_e^r$ is a bounded projection of the contextual reads; for a non-reader it is a projection of the updated private state. Lexical blocks are zero for non-readers.

The write key is $k_e^r=\tanh(W_e^k h_e^{r+1})$. With $s^r=W_s[g;b^r_{\rm ctrl}]$, routing is
\begin{equation}
\pi^r=\operatorname{softmax}\left(
\frac{(s^r)^\top k_0^r}{\sqrt{128}},\ldots,
\frac{(s^r)^\top k_3^r}{\sqrt{128}},\eta(g,b^r_{\rm ctrl})\right).
\end{equation}
The final score belongs to \NULL; its packet is zero. Source-packet interventions take place after the current write keys are computed, so they do not directly replace the current routing scores. They can alter future routing through recurrence.

For the receiver $o$, let $a_t$ combine $g$ and a projection of $h_o^4$. Vocabulary logits are
\begin{equation}
z_t=sE\,\operatorname{LN}\left[\phi(a_t)+\sigma(\lambda)W_u u_t
+\frac12\sum_{h=1}^{2}\tanh(\gamma_h(a_t))\ell^{\rm last}_{oh}\right],
\quad s=1/16.
\label{eq:fullhead}
\end{equation}
The projection is tied to the vocabulary embedding. Only the receiver's state and its own lexical reads enter this head. The PAD embedding is constrained to zero. Teacher forcing supplies earlier answer tokens during training. Algorithm~\ref{alg:token} summarizes the computation of one token.

\begin{algorithm}[h]
\caption{One native answer token in the base model}\label{alg:token}
\begin{algorithmic}[1]
\Require Independently encoded records; question anchor; receiver $o$; generated prefix
\State Compute $u_t,g$; initialize $h_e^0=0$ and $b^0=0$
\For{$r=0,1,2,3$}
\State Each module receives $b^r$; fact readers form queries and read only visible records
\State Compute coupled contextual/lexical reads and update private states
\State Construct module keys and control-plus-lexical packets
\State Compute competition with \NULL\ and pool $b^{r+1}=\sum_e\pi_e^r p_e^r$
\EndFor
\State Produce logits from the public features, receiver's state, and receiver's last lexical reads
\State Emit the next token; if generation continues, reset private/workspace states for the next token
\end{algorithmic}
\end{algorithm}

\section{Data and training}
\label{app:training}
\subsection{Paired relational worlds}
For $n$ entities, each family fixes inventories of $n$ people, objects, and locations. Binding $B$ and location $S$ are bijections. Three permutations of each relation, two record orders, six question types, and all $n$ query sides yield $3\times3\times2\times6\times n=108n$ cases per family. Four local types query a relation or its inverse; the two composition types query $S\circ B$ or its inverse. The public forms include \texttt{[bind] The holder of OBJECT is PERSON.} and \texttt{[state] PERSON is in the LOCATION.} Records contain static facts.

The generator checks that no world is duplicated. Entity strings include multi-token values, and exact match includes the complete sequence.

\subsection{Curriculum}
Each model trains in two stages. In the local stage, which lasts 2,048 updates, both fact readers can read both kinds of records, and each module receives only its own gated packet as feedback in the next round, in place of the shared broadcast. The mixed stage starts from that checkpoint with a fresh optimizer, restricts each reader to its own record type and replaces the private feedback by the shared broadcast.

\rev{The batch size is 32.} Mixed updates contain sixteen local cases and sixteen compositions, balanced across task types. The objective is
\begin{equation}
\mathcal L=\frac{1}{N_v}\sum_{t\in\mathcal V}-\log P(y_t\mid x,y_{<t})
+\frac{0.1}{N_e}\sum_{t\in\mathcal E}-\log P(y_t\mid x,y_{<t}),
\end{equation}
with value and EOS denominators computed across the batch. Optimization uses AdamW \citep{loshchilov2019}, $(\beta_1,\beta_2)=(0.9,0.95)$, $\epsilon=10^{-8}$, weight decay 0.01, norm clipping at one, and FP32 arithmetic. The mixed-stage peak learning rate is $10^{-4}$, with sixteen warmup updates and cosine decay to $10^{-5}$, shown here for a mixed stage of 2,048 updates:
\begin{equation}
\operatorname{lr}(u)=
\begin{cases}
10^{-4}u/16,&u\le16,\\
10^{-4}\left[0.1+0.9\,\dfrac{1+\cos(\pi(u-16)/2032)}2\right],&u>16.
\end{cases}
\end{equation}
The local stage uses a peak of $3\times10^{-4}$: a cosine schedule over its first 256 updates (cosine denominator 496), followed by continuous decay to $3\times10^{-5}$.

The main models train on 32 families per corpus for 12,288 updates (Appendix~\ref{app:exp-main}), and the confirmation models on 8 families for 4,096 updates (Appendix~\ref{app:g4}), which takes about 21 minutes per model with under 1\,GB of allocated tensor memory. Training lengths are fixed in advance and every result uses its complete panel, so no checkpoint or subset is selected by its intervention effect.

\FloatBarrier
\section{The confirmation models: frozen confirmation and further analyses}
\label{app:g4}
\rev{This appendix collects everything measured on the confirmation models (A1, A2, B1 and B2). They have the architecture, corpora and evaluation panels of the main models and were trained on 8 families per corpus for 4,096 updates, under a plan fixed before training. Sections~\ref{app:results-ext} and~\ref{app:g4-results} report the frozen confirmation. Sections~\ref{app:exp-off}--\ref{app:exp-relayarms} report further analyses of these models. A \emph{lineage} is one confirmation model together with the variants trained from its local-stage checkpoint. The main models are reported in Appendix~\ref{app:exp-main}.}
\subsection{Results by rung}
\label{app:results-ext}
Table~\ref{tab:g4-summary} gives every primary count for every model.

\begin{table}[!htbp]
\caption{The confirmation models. Accuracy is complete-answer exact match including EOS. The workspace on/off rows and all rows below them use 512 cases per model. Restoration, query noise and installation are defined in Section~\ref{sec:mechanism}. Changed cases are those in which the transplant changes the native output.}
\label{tab:g4-summary}
\centering\small
\begin{tabular}{@{}lrrrr@{}}
\toprule
& A1 & A2 & B1 & B2\\
\midrule
Composition accuracy on the held-out split (of 2,304) & 1,499 & 1,475 & 1,651 & 1,739\\
Composition correct, workspace on & 332 & 332 & 360 & 404\\
Composition correct, workspace off & 117 & 120 & 129 & 140\\
Local correct, workspace on & 417 & 394 & 432 & 446\\
Local correct, workspace off & 410 & 390 & 430 & 441\\
\midrule
Hybrid answers $Z$ without intervention & 65 & 56 & 44 & 25\\
Hybrid answers $Z$ after the transplant & 312 & 317 & 368 & 374\\
Hybrid answers $Z$ with restoration & 65 & 56 & 44 & 25\\
Hybrid answers $Z$ with query noise & 303 & 309 & 359 & 371\\
Hybrid answers $Z$ with installation alone & 311 & 317 & 367 & 373\\
\midrule
Transplant output = the model's output on the hybrid world & 469 & 487 & 490 & 462\\
Restoration output = native output & 512 & 512 & 512 & 511\\
Both read tests hold, of changed cases & 356/357 & 377/377 & 397/398 & 421/423\\
\bottomrule
\end{tabular}
\end{table}

\textbf{Rung 1: dependence.} All four models fit all 3,456 training questions. On held-out worlds, composition accuracy is 64--75\% and local accuracy 79--86\%. Switching the workspace off lowers composition accuracy by 41--52 points and local accuracy by 0.4--1.4 points. The loss is not a net of offsetting changes: 243--277 composition answers per model flip from correct to wrong, against 13--31 in the other direction.

\textbf{Rung 2: content.} The write-2 transplant yields the hybrid answer $Z$ in 61--73\% of cases, against 5--13\% without intervention. The transplanted output equals the same model's native output on the hybrid world in 90--96\% of cases, errors included, so the receiver behaves as though it had been given the hybrid world, and most misses reproduce the model's own error on that world (Appendix~\ref{app:analysis-fidelity}). On secondary panels of 128 cases, self-replacement and a transplant at the never-read write 3 leave every output unchanged. Earlier writes have smaller effects (16--27 hybrid answers at write 0 and 31--41 at write 1, against 81--96 for writes 0--2), as expected when the final read consumes content written late. A donor with the same facts in a different record order changes 13--32 outputs, which Appendix~\ref{app:exp-otherarch} traces to the order-sensitive reader.

\begin{figure}[!htbp]
\centering\includegraphics[width=\linewidth]{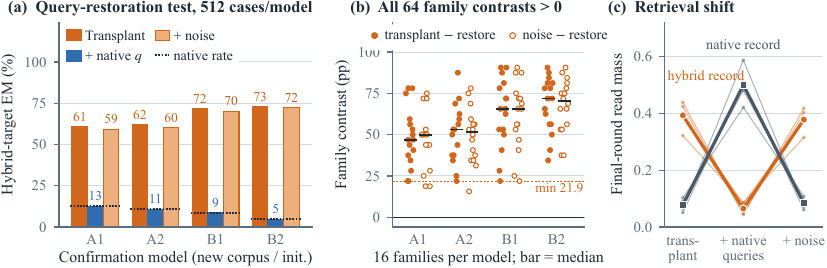}
\caption{\textbf{The retrieval route in the confirmation models.} (a)~The source packet is transplanted at write $r=2$; the other two conditions additionally restore, or perturb with matched norm, the receiver's $r=3$ queries. Each bar covers all 512 preselected cases per model; numbers are percentages, and dotted lines show the native rate of hybrid answers. (b)~Family-level differences in hybrid rate, one point per family (32 related recipients); bars mark medians and the dashed line the smallest transplant-minus-restoration difference. (c)~Final-round read mass on the record needed for the hybrid answer and on the native record; thin lines are models, thick lines their mean.}
\label{fig:g4-query}
\end{figure}

\FloatBarrier
\textbf{Rung 3: mechanism.} With the transplant in place, restoring the receiver's native queries returns the number of hybrid answers exactly to its native value in every model and recovers the complete native output in 2,047 of 2,048 cases (Figure~\ref{fig:g4-query}a). Query noise of the same norm keeps 97--99\% of the effect. Installing only the transplant-induced queries gives a hybrid rate within one case of the transplant's in every model, and both read tests hold in 1,551 of the 1,555 cases the transplant changes (Appendix~\ref{app:analysis-factorial}). The four remaining cases are retained in every count. The final read follows the queries: its mass on the record needed for $Z$ is 0.32--0.44 under the transplant and 0.04--0.09 under restoration (Figure~\ref{fig:g4-query}c). The effect holds in every family. The hybrid rate under the transplant exceeds that under restoration in all 64 family--initialization combinations, by at least 21.9 points (Figure~\ref{fig:g4-query}b), and a one-sided sign test over the 32 families gives $p = 2^{-32} < 10^{-9}$.

\subsection{Plan, data and training}
\label{app:g4-results}
Before any confirmation training, we fixed two data-generator seeds and four model seeds, the 6,891,784-parameter architecture, the two-stage curriculum and its learning rates, the training length, the primary intervention site (write 2 and read 3), and the membership of all panels. No evaluation of a confirmation model was used to change any of them.

Corpus A uses generator seed 20260923401, with model seeds 20260923411 (A1) and 20260923412 (A2). Corpus B uses generator seed 20260923402, with model seeds 20260923421 (B1) and 20260923422 (B2). Each corpus contains 8 training, 4 development, and 16 confirmation families of 432 questions each. No complete fact world is shared across the six corpus and split combinations, so the confirmation split tests recombination in new worlds under the same generating setup. Each model trains its local-stage ancestor for 2,048 updates and its mixed-stage branch for another 2,048 with a fresh optimizer and batch size 32, as described in Appendix~\ref{app:training}.

The primary panel selects 16 eligible cases per composition direction within each of the 16 confirmation families, 512 recipients per model. Eligibility is defined by the symbolic distinction of $X$, $Y$ and $Z$, never by a model output, and the two initializations of a corpus use the same panel. Secondary write sites and controls each use 128 predetermined cases.

\subsection{Rung 1: ablation method and the memory-only reference}
\label{app:exp-off}
We re-ran the four confirmation models on their original OFF panels. Besides zero-OFF (Definition~\ref{def:off}), we applied mean ablation (a task-matched mean broadcast from different worlds, per token position and round) and resample ablation (the whole broadcast of a random different-world case of the same task; three draws). Table~\ref{tab:exp-off} gives composition accuracy together with the memory-only reference.

\begin{table}[!htbp]
\caption{Composition OFF panels (512 cases per lineage). Memory-only: expected panel score of the Bayes-optimal predictor that sees only the receiver's memory. Resample: mean of three draws.}
\label{tab:exp-off}
\centering\small
\begin{tabular}{@{}lrrrrr@{}}
\toprule
Lineage & Native & Memory-only & Zero-OFF & Mean-OFF & Resample-OFF\\
\midrule
A1 & 332 & 271.0 & 117 & 115 & 114.0\\
A2 & 332 & 271.0 & 120 & 120 & 108.3\\
B1 & 360 & 238.0 & 129 & 131 & 120.0\\
B2 & 404 & 238.0 & 140 & 118 & 114.3\\
\bottomrule
\end{tabular}
\end{table}

\textbf{Findings.} (i) Every OFF method leaves 108--140 correct, 98--163 cases below the memory-only reference; under OFF the models fall far short of what their own memory would allow. (ii) By exact match, the composition deficit holds under every ablation method in all 8 runs (four lineages, on the original panels and on unseen panels), each costing 205--293 cases. By teacher-forced sequence log-probability it holds for every method that ablates the whole answer. Mean ablation of the first answer token alone lowers the model's confidence without changing its answer, and shows the deficit in 3 of 8 runs. (iii) On local questions, zero and mean ablation change accuracy by $-14$ to $+17$ cases, whereas resample ablation, which inserts a foreign broadcast, costs 8--76 cases. Zero-OFF remains the primary method because its outputs are constant within each information class, as Proposition~\ref{prop:off} requires (0 violations in every run). (iv) The rung-2 verdict does not depend on the metric: exact match and teacher-forced sequence log-probability agree in all 32 comparisons (eight runs, single-packet and trajectory transplants, three-value and two-value panels).

\textbf{Site control and a probe of the pre-read state.} Restoring the receiver's round-2 queries instead of its round-3 queries returns 0--3 of the trajectory transplant's changed cases to native, against 99.3--100\% for round-3 restoration (site-specific in 8 of 8 runs). A ridge probe does not decode the receiver's alternative person$\to$place assignments from the pre-read state $h^3$ beyond what the question cue alone gives (35.1--36.1\% with $h^3$, against 34.7--35.0\% for the cue alone; chance 33.3\%), although the same probe decodes the model's own answer from the round-3 read with 94.0--98.5\% accuracy (positive control). The state $h^3$ carries the native answer (57--71\% for the true $X$). The causal test of route (b) is the decomposition of Appendix~\ref{app:analysis-factorial}, in which the broadcast changes no output once the native queries are restored.

\subsection{Rung 2: relay on actual donor outputs, and two-value patching}
\label{app:exp-relay}
The symbolic donor answer $Y$ differs from $Z$ by construction, but a model can answer the donor input wrongly, so relay is scored on the donor's actual output. The informative cases are those in which this output is neither the native output nor $Z$, where relay and recombination predict different strings. For the main models the transplant copies the donor in 12 of 1,994 such cases and gives $Z$ in 1,969 (Appendix~\ref{app:exp-main}). For the confirmation models it copies the donor in 99 of 1,257 and gives $Z$ in 1,050. It does not raise outputs of the donor's symbolic answer $Y$ above the native rate (202 against 204 of 2,048 on the unseen panels), and the receiver's read mass on the record holding the donor's value stays at or below 0.10 in every transplant condition.
\textbf{Two-value patching.} With a standard donor, whose answer is also the hybrid answer, the same transplant ``follows the donor'' in 316--380 of 512 cases per run, the same headline as the three-value panel. Relay and recombination predict the same string in every such case, so the two-value design cannot separate them.

Appendix~\ref{app:exp-relayarms} scores the three-value verdict blind against lesion ground truth in trained variants in which the source can read the receiver's memory, so that relay is available.

\subsection{Rungs 2 and 3 on unseen panels, and a third-entity control}
\label{app:exp-fresh}
To repeat the $2\times2$ design on unseen data, we generated two new panels, one per corpus, with the unchanged data generator of the confirmation study. Each holds 512 three-value cases from 16 families, and no complete world and no entity set is shared with any earlier training, development or confirmation data, or between the two panels. The panels and the evaluation plan were fixed before the first forward pass. The two lineages of a corpus are evaluated on the same panel, so the 2,048 cases are model--case pairs over 1,024 distinct recipients. We reran the $2\times2$ design unchanged on the four confirmation models (Table~\ref{tab:exp-fresh}).

\begin{table}[!htbp]
\caption{The $2\times2$ design on unseen panels (512 three-value cases per lineage). Changed: the transplant changes the native output. Restore: restoring the receiver's round-3 queries gives the native output. Query-only: installing only the transplant-induced queries reproduces the transplant output. Joint: both hold, among changed cases.}
\label{tab:exp-fresh}
\centering\small
\begin{tabular}{@{}lrrrrrrr@{}}
\toprule
Lineage & Native $X$ & Accidental $Z$ & Transplant $Z$ & Changed & Restore & Query-only & Joint\\
\midrule
A1 & 352 & 46 & 332 & 392 & 512/512 & 508/512 & 388/392\\
A2 & 333 & 53 & 332 & 391 & 512/512 & 512/512 & 391/391\\
B1 & 364 & 64 & 360 & 392 & 512/512 & 512/512 & 392/392\\
B2 & 359 & 64 & 341 & 359 & 512/512 & 512/512 & 359/359\\
\midrule
Total & 1,408 & 227 & 1,365 & 1,534 & 2,048/2,048 & 2,044/2,048 & 1,530/1,534\\
\bottomrule
\end{tabular}
\end{table}

The unseen panels reproduce the confirmation results. The transplant yields $Z$ in 1,365 of 2,048 cases, against 227 by accident. Restoring the receiver's native queries returns every output to native, and installing only the induced queries reproduces the full transplant in 2,044 cases. The four exceptions all come from lineage A1 and are retained. The transplanted recipient matches the model's own native output on the hybrid world in 1,920 of 2,048 cases.

A redirected query could in principle act as a generic disturbance that happens to land on $Z$. The third-entity control rules this out. For forward-composition cases we chose a second donor that binds the queried object to a third person $C$, different from the people in the recipient and the primary donor, such that $X$, $Z$, the third-entity hybrid answer $Z_C$ and that donor's answer are distinct. With the receiver's memory unchanged, installing only the round-3 queries that this donor's packet induces yields $Z_C$ in 200 of 272 eligible model--case pairs (53/74, 47/74, 53/62 and 47/62 for A1, A2, B1 and B2), against 24 of 272 natively, and it reproduces the full third-entity transplant's output in 272 of 272. The receiver's queries therefore carry which person to look up, not merely a change of query.

\textbf{Larger worlds.} On the eight-entity panels of Appendix~\ref{app:exp-scale}, the confirmation models answer 508 of 1,024 composition questions without retraining, and OFF lowers this to 120. The transplant raises hybrid answers from 77 to 513, about as many as the models answer correctly natively. Restoration returns 1,020 of 1,024 outputs to native, query-only installation reproduces the transplant in 1,021, and both hold in 798 of 804 changed cases.

\subsection{Routing at the transplant site}
\label{app:exp-routing}
\textbf{(a) Routing weights in the confirmation models.} At write 2 of the first answer token, the router puts mean weight 0.956--0.995 on the source module in every lineage and composition direction, and the source has the largest weight in every case (0/256 exceptions per lineage and direction). Scaling the source's weight to $w$ while transplanting scales the effect: the share of the transplant effect obtained is 0.01--0.03 at $w=0.01$, 0.46--0.60 at $w=0.3$, and 1.00--1.02 at $w=0.97$. Setting the source's weight to zero at write 2 lowers native composition accuracy from 333--364 to 198--236 of 512 when the weight is moved to NULL, and to 168--266 when the other weights are renormalized instead. Routing weights at write 2 vary little across inputs: their mean total-variation distance between recipient and donor (0.018--0.034) is below that between two unrelated recipients of the same task (0.023--0.038).

\textbf{(b) Trained uniform-routing baseline (plan fixed in advance).} For each lineage we trained a uniform-routing model whose routing weights are fixed at $1/5$ on all five slots (four modules and NULL). It runs the mixed stage (same data order, schedule, and 2,048 updates) from the lineage's own local-stage checkpoint. The local stage has no inter-module competition, so the uniform-routing model never experiences competition. The analysis and its reading were fixed before any uniform-routing checkpoint was evaluated.

\begin{table}[!htbp]
\caption{Learned routing (L, the confirmation models) versus trained uniform routing (U), 512 three-value cases per lineage. OFF deficit: native $X$ minus zero-OFF $X$. Transplant: single write-2 packet, with the accidental (native) $Z$ count in parentheses. Restore: transplant-changed cases returned to native by round-3 query restoration. Trajectory: writes 0--2.}
\label{tab:exp-uniform}
\centering\small
\begin{tabular}{@{}lrrrrrrrrrr@{}}
\toprule
& \multicolumn{2}{c}{Native $X$} & \multicolumn{2}{c}{OFF deficit} & \multicolumn{2}{c}{Transplant $Z$} & \multicolumn{2}{c}{Restore} & \multicolumn{2}{c}{Trajectory $Z$}\\
\cmidrule(lr){2-3}\cmidrule(lr){4-5}\cmidrule(lr){6-7}\cmidrule(lr){8-9}\cmidrule(l){10-11}
Lineage & L & U & L & U & L & U & L & U & L & U\\
\midrule
A1 & 352 & 351 & 222 & 222 & 332 (46) & 250 (44) & 392/392 & 309/309 & 342 & 341\\
A2 & 333 & 333 & 217 & 227 & 332 (53) & 308 (49) & 391/391 & 358/358 & 345 & 349\\
B1 & 364 & 370 & 241 & 245 & 360 (64) & 342 (54) & 392/392 & 373/373 & 361 & 365\\
B2 & 359 & 361 & 216 & 208 & 341 (64) & 265 (63) & 359/359 & 275/275 & 346 & 352\\
\bottomrule
\end{tabular}
\end{table}

\textbf{Result.} All four lineages retain the signature (Table~\ref{tab:exp-uniform}). Capability and the OFF deficit match learned routing, the transplant exceeds the accidental rate by 202--288 cases, restoration returns every changed case to native, and query-only reproduces the transplant in 1,312 of 1,315 changed cases. Broadcast use therefore holds with and without learned competition (Table~\ref{tab:commitments}).

Learned routing concentrates the content in one packet. Under uniform routing the single write-2 packet yields $Z$ less often in every lineage, whereas the transplant of writes 0--2 yields $Z$ as often as under learned routing (341--365 against 342--361 of 512). This is what fixed weights predict: the source's write-2 packet becomes one of five equally weighted terms instead of carrying a learned weight of 0.956--0.995.

\subsection{Relay-capable variants, scored blind against lesion ground truth}
\label{app:exp-relayarms}
\textbf{Design.} In the confirmation models the source never reads the receiver's memory (typed visibility), so it has no direct access to the facts that the donor's answer requires. We trained variants that change only what each module may read on composition rows, each on the four lineages: one mixed stage of 2,048 updates from the lineage's local-stage ancestor with the original recipe. In \emph{FREE} the source reads both memories, so relay is available but not rewarded. \emph{RECOMB} is FREE with the source's view of the receiver-type records resampled at random during training, so only recombination solves the task. In \emph{SELF} the receiver reads both memories, so it could compose without the workspace. \rev{FREE and SELF are the variants that Appendix~\ref{app:exp-whocomputes} calls ``source reads both memories'' and ``receiver reads both memories'' on the main models.} A resampled record keeps its slot and person and draws its answer uniformly from the world's four candidates. Local rows are never altered, and labels are always true.

\textbf{Ground truth and blinding.} Independently of the protocol, inference-time lesions on the 512 three-value recipients show which memory each checkpoint's answers come from. We resample the receiver's own records, the source's view of them, or the workspace (three draws each). A model \emph{recombines} if its output follows the resampled receiver record in at least half of the informative cases while the source lesion breaks fewer than 20\%, and it \emph{relays} in the mirror case. The three-value verdict is \emph{relay} if most informative outputs copy the donor's actual output and \emph{recombination} if most are $Z$. Both rules were fixed before any forward pass on these checkpoints. The checkpoints (the variants and the four confirmation models) were coded at random, and both rules ran on the coded outputs before unblinding.

\begin{table}[!htbp]
\caption{What each check concludes, per arm (ranges over 4 lineages; 512 three-value recipients per checkpoint). AFGW denotes the four confirmation models. Informative: cases in which the donor's actual output is neither native nor $Z$ and the transplant changes the output. Lesion columns count (correct case, draw) pairs over three draws: for the receiver and source lesions, pairs whose resampled record differs from the truth (source breaks are out of the same pairs); for the workspace lesion, pairs whose donor answer differs.}
\label{tab:exp-relayarms}
\centering\scriptsize
\setlength{\tabcolsep}{2.5pt}
\begin{tabular}{@{}lccccccccc@{}}
\toprule
 & native & zero-OFF & \multicolumn{3}{c}{three-value transplant (S, write 2)} & \multicolumn{3}{c}{lesion ground truth} & verdict\\
\cmidrule(lr){4-6}\cmidrule(lr){7-9}
Arm & comp. & comp. & informative & copies donor & $Z$ & follows receiver & source breaks & workspace breaks & = truth\\
\midrule
AFGW & 333--364 & 116--143 & 285--313 & 15--30 & 251--265 & 722--783 / 744--821 & 0 & 653--704 / 861--936 & 4/4\\
FREE & 338--365 & 121--165 & 290--315 & 19--26 & 236--272 & 722--792 / 751--824 & 1--6 & 653--705 / 873--940 & 4/4\\
RECOMB & 343--367 & 119--148 & 304--326 & 19--29 & 248--273 & 728--794 / 764--827 & 1--2 & 650--710 / 883--951 & 4/4\\
SELF & 335--364 & 113--138 & 299--309 & 21--31 & 241--267 & 713--787 / 744--828 & 0 & 662--692 / 865--941 & 4/4\\
\bottomrule
\end{tabular}
\end{table}

\textbf{When relay is available, all four lineages still recombine.} By lesion ground truth (Table~\ref{tab:exp-relayarms}), FREE computes its answers from the receiver's memory in 4/4 lineages: under the receiver lesion the output follows the resampled record in 722--792 of 751--824 informative pairs, while the source lesion breaks 1--6. RECOMB and SELF do the same, and so do the confirmation models, for which the source lesion is an exact no-op. The three-value verdict says recombination for every one of these 16 checkpoints, on both the single-packet and the trajectory transplant, and so agrees with the blind ground truth in 16/16. Because all 16 recombine, this agreement shows that the check does not call relay where there is none. Appendix~\ref{app:exp-reference} shows the complementary property, that it reports relay where relay occurs. OFF (zero, mean and resample) shows a deficit and the two-value transplant shows content transfer (356--404 of 394--426 eligible cases) in all 16, as in the main results. Read-3 restoration returns 6,159 of the 6,164 changed cases to native.

\textbf{A receiver that could bypass the workspace still uses it.} Zero-OFF costs 214--246 of 512 compositions, the transplant gives $Z$ in 241--267 of 299--309 informative cases, and the workspace lesion breaks 662--692 of 865--941 correct answers. Although its receiver can read both memories, training from the local-stage ancestor still makes it key its own memory on the broadcast person, as the confirmation models do. Use of the workspace is thus learned even where the architecture makes it optional.

\section{Intervention definitions and metrics}
\label{app:interventions}
\subsection{Extended protocol description}
\label{app:protocol-ext}
\rev{\textbf{Proof of Proposition~\ref{prop:off}.} Fix a generated token $t$ and apply OFF (Definition~\ref{def:off}): every module receives $b^r=0$ in every round. The receiver's feedback input $f_o^r=W_o^f[h_o^r;g;0]$ and query $q_o^r$ then depend only on $(h_o^r,u_t,g)$. Its reads depend only on that query and on the records of its own bank $\mathcal{B}_o$, each of which is encoded independently of every other record. Its update $h_o^{r+1}$ depends only on $h_o^r$, $u_t$, $g$, $f_o^r$, and those reads. By induction over $r$ from $h_o^0=0$, both $h_o^4$ and the receiver's final lexical reads are functions of $(u_t,g,\mathcal{B}_o)$. The other modules' states, packets, and routing weights are still computed, but under OFF they reach the receiver only through broadcasts that are zeroed on receipt. The head uses only $g$, $h_o^4$, $u_t$, and the receiver's final lexical reads, and greedy decoding is deterministic, so token $t$ is a function of $(u_t,g,\mathcal{B}_o)$. Here $g$ depends only on the question, and $u_t$ depends only on the question and the tokens generated before $t$. Induction over tokens, with the receiver fixed by the question type, shows that the complete sequence, including EOS, is a function of the question and $\mathcal{B}_o$. \qed

\begin{corollary}[Memory-only ceiling]\label{cor:ceiling}
Partition any panel (fixed evaluation set) of $n$ cases with answers $y_i$ into classes $\kappa$ of cases with the same question and the same $\mathcal{B}_o$. OFF answers all cases of a class identically, so OFF accuracy is at most the memory-only ceiling $\frac1n\sum_\kappa\max_a|\{i\in\kappa:y_i=a\}|$. Over a distribution of inputs, it is at most the accuracy of the Bayes-optimal predictor that sees only the question and $\mathcal{B}_o$.
\end{corollary}

\textbf{Proof of Corollary~\ref{cor:ceiling}.} Within a class of cases with the same question and the same $\mathcal{B}_o$, OFF produces one output, which is correct for at most the cases whose answer equals it; the best case is the most frequent answer of the class. Summing over classes gives the panel bound. For a distribution, the same argument conditional on the question and $\mathcal{B}_o$ gives at most $\max_a P(y=a\mid \text{question},\mathcal{B}_o)$, whose expectation is the accuracy of the Bayes-optimal receiver-memory-only predictor. \qed

\textbf{How far the models fall below it.} As the memory-only reference we use the expected score of the Bayes-optimal predictor that sees only the question and the receiver's memory, which is 271.0 (corpus A) and 238.0 (corpus B) of the 512 composition cases. The main models answer 490--508 of these cases natively and 108--137 with the workspace off (21--27\%, against 46--53\% for the reference), 119--147 cases below the reference (Appendix~\ref{app:exp-main}). The confirmation models answer 332--404 natively and 117--140 with the workspace off, 98--154 cases below it (Appendix~\ref{app:exp-off}). For local questions the receiver's memory determines the answer, and local accuracy barely changes under OFF. The trained models thus route composition information through the workspace, and they depend on it beyond what the architecture requires.}

\begin{figure}[!htbp]
\centering\includegraphics[width=\linewidth]{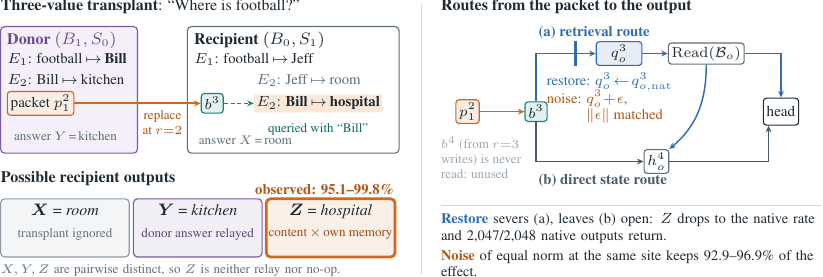}
\caption{\textbf{The content and mechanism tests on a real case from the primary panel.} Left: the donor's write-2 packet from the binding reader replaces the recipient's. Room ($X$), kitchen ($Y$), and hospital ($Z$) indicate an ignored transplant, relay, and recombination. Right: the packet reaches the output only through $b^3$, either via the receiver's round-3 queries and reads (route a) or directly via its final state (route b). Restoration closes route (a) and leaves route (b) open.}
\label{fig:intervention}
\end{figure}

\textbf{Routes.} Figure~\ref{fig:intervention} traces the running example of the main text, a real case from the primary panel, through the model. In the primary intervention the transplanted packet enters the computation only through $b^3$, which reaches the output by two routes. Along route (a), $b^3$ changes the receiver's round-3 queries and hence its reads, which enter the output head and the receiver's final state. Along route (b), $b^3$ enters the receiver's final state directly. The other modules' round-3 updates only produce $b^4$, which nothing reads.

\textbf{What the controls match.} Contextual and lexical reads share their queries, so the read tests address them together. Query noise matches the size of the query change that restoration makes, measured by its Euclidean norm. The third-entity control of Appendix~\ref{app:exp-fresh} matches its kind: a query redirected to a different person yields that person's answer.

\subsection{Worlds, packet sites, and exact targets}
For a forward composition, write the donor world $(B_1,S_0)$, the recipient world $(B_0,S_1)$ and the hybrid world $(B_1,S_1)$, with answers $Y$, $X$ and $Z$. The primary panel requires the three to be pairwise distinct. For a reverse composition the two relations exchange roles. One symbolic interpreter computes every target from the records.

The transplant acts on the source module on the first answer token, at one or more writes in $\{0,1,2\}$. It replaces the entire 640-dimensional packet, after key construction and before pooling, with the donor's own activation. Other packets and the current routing scores are left as they are, later states are recomputed, and generation proceeds without intervention after the first answer token.

\subsection{Native-query restoration and its noise control}
Restoration overwrites the receiver's read queries at every round after the earliest intervened write with the queries recorded on the same recipient input without intervention:
\begin{equation}
q_{oh}^k \leftarrow q_{oh,\mathrm{native}}^k, \qquad k > \min \mathcal{W}, \quad h \in \{1,2\},
\label{eq:restore}
\end{equation}
where $\mathcal{W}$ is the set of intervened writes. For the primary site ($\mathcal{W}=\{2\}$), only the final read at $k=3$ is restored. Only the queries are overwritten. Private states, records, source packets and the output head are recomputed.

The noise condition keeps the transplant and moves the receiver's query $\widetilde q^k_h$ along a seeded random unit direction $\widehat\epsilon^k_h$ by the distance that restoration moves it:
\begin{equation}
q^k_{h,\rm noise}=\widetilde q^k_h+
\|q^k_{h,\rm native}-a^k_h\|_2\,\widehat\epsilon^k_h,
\end{equation}
where $a^k_h$ is the query that restoration replaces. The distance is measured for each example, round and head.

\subsection{Content and implementation controls}
Five controls accompany the transplant. Self-replacement shows that the intervention itself changes nothing. A donor that holds the same facts in a different record order keeps the relations and changes the ordered reader's input. An irrelevant-source change keeps the facts that a local question needs and alters the other memory. A transplant at write 3 tests a site that nothing reads. Packet noise replaces the difference between the donor's and the recipient's packet with random directions of the same norm, separately in the control block and in each lexical block. Block-isolation conditions transplant only the control block or only the lexical blocks (Appendix~\ref{app:exp-main}).

\subsection{Metrics}
All quantities use the full preselected panel, including cases that the model answers wrongly without intervention. The hybrid rate is exact match to $Z$ followed by EOS. Hybrid agreement is identity with the same model's native output on the hybrid world, errors included, and asks whether the receiver behaves as if it had been given the hybrid world. Native recovery is identity with the unperturbed output. Read mass is the weight that a read places on the record that the native or the hybrid answer requires, with the \NULL\ mass kept in the total.

\section{Three-receiver variant}
\label{app:reuse}
\subsection{Design and evaluation}
\label{app:reuse-main}
GWT's broadcast is meant to be globally available. To test this property directly, we trained a separate variant with one source module that reads object $\to$ person bindings and three receivers that read person $\to$ location, person $\to$ color, and person $\to$ access mappings. One transplanted source packet therefore implies a different hybrid answer at each receiver. We trained two initializations (seeds 106101 and 106102) with answer-only loss.

\rev{\textbf{Evaluation.} Rung 4 is evaluated with the source's single write-2 packet, donors whose maps differ from the recipient's, a panel of 512 joint episodes from four new families fixed in advance, and rung 3 per receiver (Section~\ref{sec:reuse}; full results in Appendix~\ref{app:exp-breadth}). With such donors the donor's answers differ from the hybrid answers, so the test separates relay from recombination at each receiver. A separate selectivity test transplants the source's whole write trajectory with donors that share the recipient's maps: changing one receiver's mapping, or cutting one receiver's input, leaves the other two receivers' outputs unchanged (zero changes in 1,024 comparisons per condition and seed).}

\subsection{Architecture, decoding schedule, and training}
Each module reads its own typed memory. Four rounds process the shared question, and each receiver obtains private cached answer features and lexical reads, from which its later tokens are decoded, so no receiver's answer tokens re-enter the shared rounds or reach another receiver. Each family uses four entities, two deranged choices for each of the four relation maps, two record orders, and four query sides, which gives $2^4\times2\times4=128$ joint episodes per family. A joint episode requires three complete answers, and eight training families yield 1,024 joint episodes. Local and mixed stages each use 2,048 updates with \rev{batch size 32} and final-answer losses, and the mixed stage combines sixteen local questions with sixteen joint episodes. \rev{The evaluation transplants write 2 alone, writes 0 and 1 alone as controls, and the trajectory of writes 0--2. Write 3 is never read.}

\FloatBarrier
\setcounter{section}{7}
\section{Analytical properties of the intervention design and the query-only test}
\label{app:analysis}

\subsection{Fidelity versus hybrid rate}
\label{app:analysis-fidelity}
A transplant can miss the hybrid answer $Z$ for two reasons. The content may not have been used, or it was used and the model errs on the hybrid world as it does without any intervention. The two are told apart by comparing the transplanted output with the same model's native output on the hybrid world. If $t$ transplanted outputs equal $Z$ and $a$ equal the model's hybrid-world output, then at least $a-t$ of these agreements are errors that the model also makes natively. For the confirmation models (Table~\ref{tab:g4-summary}) this accounts for at least 537 of the 677 misses (157, 170, 122 and 88 per model), so the hybrid rate there is limited mainly by task accuracy. For the main models the transplanted output equals the hybrid-world output in 2,021 of 2,048 cases (Appendix~\ref{app:exp-main}).

\Needspace{17\baselineskip}
\subsection{First-token equivalence under per-token resets}
\label{app:analysis-reset}
\begin{proposition}[First-token equivalence]
\label{prop:analysis-reset}
Assume deterministic greedy decoding, the base model's per-token reset of all private and workspace states, unchanged input memories, and an intervention $I$ that acts only during the first generated token. Then the first token after the intervention equals the native first token if and only if the complete generated sequences, including EOS, are equal.
\end{proposition}
\begin{proof}
If the first tokens agree, the next computation has the same prefix, question and memories. The reset states are identical and no intervention remains, so the next token is identical. Induction over tokens gives the same sequence. The converse is immediate.
\end{proof}

Rungs 2 and 3 intervene on the first answer token only, so a recovered first token is a recovered complete sequence. The three-receiver variant decodes from private caches and is evaluated on complete answers directly (Appendix~\ref{app:reuse}).

\subsection{Necessity, sufficiency, and interaction}
\label{app:analysis-factorial}
Consider the first answer token and the write at $r=2$, followed by the receiver's read at $r=3$. Let $(b_{\mathrm{native}},q_{\mathrm{native}})$ be the native broadcast and queries of the recipient and $(b_{\mathrm{trans}},q_{\mathrm{trans}})$ the values that the transplant produces in that same recipient. The receiver's two queries are taken together. Table~\ref{tab:analysis-factorial} gives the four conditions.

\begin{table}[!htbp]
\centering\small
\caption{The four conditions at the receiver's final read. The index gives the broadcast and the queries (0 native, 1 transplanted).}
\label{tab:analysis-factorial}
\begin{tabular}{@{}llll@{}}
\toprule
Index & Condition & Broadcast & Retrieval queries\\
\midrule
00 & Native & $b_{\mathrm{native}}$ & $q_{\mathrm{native}}$\\
11 & Transplant & $b_{\mathrm{trans}}$ & $q_{\mathrm{trans}}$\\
10 & Restoration & $b_{\mathrm{trans}}$ & $q_{\mathrm{native}}$\\
01 & Installation (query-only transplant) & $b_{\mathrm{native}}$ & $q_{\mathrm{trans}}$\\
\bottomrule
\end{tabular}
\end{table}

Installation and restoration override only the receiver's queries. The private state, the memory records, the output head and all later steps are recomputed, and both contextual and lexical retrieval run on the overridden queries.

\begin{proposition}[Exact factorial decomposition]
\label{prop:analysis-factorial}
For any scalar response $m_{ij}$ of these four runs,
\begin{equation}
m_{11}-m_{00}
=(m_{01}-m_{00})+(m_{10}-m_{00})+(m_{11}-m_{10}-m_{01}+m_{00}).
\label{eq:analysis-factorial}
\end{equation}
The three terms are the change due to the queries alone, the change due to the broadcast at the native queries, and their interaction.
\end{proposition}

The identity holds by cancellation. It shows why two tests are needed. Restoration measures the second term and so tests whether redirected retrieval is necessary. Installation measures the first term and so tests whether it is sufficient. A response such as $m(b,q)=bq$ would pass under restoration and fail under installation.

\textbf{Results.} For the main models, restoration returns the native output in 2,047 of 2,048 cases, installation reproduces the transplant's output in 2,047, and both hold in 2,013 of the 2,015 cases the transplant changes (Table~\ref{tab:exp-main}). For the confirmation models (Table~\ref{tab:g4-summary}), installation yields 311, 317, 367 and 373 hybrid answers of 512, against 312, 317, 368 and 374 for the transplant, and both tests hold in 1,551 of 1,555 changed cases. In the decomposition of the hybrid rate, the term due to the broadcast at the native queries is zero in every model, and the interaction is at most one case in 512. Redirected queries are therefore necessary and sufficient for the change of output. Appendix~\ref{app:exp-fresh} repeats the design on unseen panels.

\Needspace{10\baselineskip}
\subsection{Order sensitivity}
\label{app:analysis-order}
AFGW's reader favors the last matching record. With a gate $m_a\in[0,1]$ for record $a$ and its token-weighted value $\bar V_a$, the read is $R=\sum_a w_a\bar V_a$ with $w_a=m_a\prod_{j>a}(1-m_j)$. Swapping two adjacent records $a$ and $b$ changes the read by
\begin{equation}
R_{\mathrm{swap}}-R_{\mathrm{original}}
=T\,m_am_b(\bar V_a-\bar V_b),\qquad T=\prod_{j\text{ after both}}(1-m_j),
\label{eq:analysis-swap}
\end{equation}
and leaves all other record weights and the NULL mass unchanged. Reordering therefore matters only when two records both match the query and hold different values. A reader that weights records by a softmax over record scores, with a fixed NULL option, is invariant to record order by construction. Appendix~\ref{app:exp-otherarch} trains such a reader, and the mechanism is unchanged.

\FloatBarrier
\section{Experiments on the main models, their variants and other systems}
\label{app:exp}

This appendix reports the main models, their variants and the other systems of the main text, in the order of Section~\ref{sec:results}. Table~\ref{tab:exp-status} in Appendix~\ref{app:guide} lists the question each experiment answers. Analyses of the confirmation models are in Appendix~\ref{app:g4}. Where an experiment's panels and analysis were fixed before its checkpoints were evaluated, its section says so. Models that share a corpus share its data order, and every count is given as $k/n$.
\subsection{Main models: training and complete results}
\label{app:exp-main}
\textbf{Models.} The main models are A3 and A4 (corpus A) and B3 and B4 (corpus B). Their architecture is that of the confirmation models, unchanged. Training adds 24 families per corpus, drawn from the same word pools, to the eight training families of the confirmation models, and runs for 12,288 updates, against 4,096 for the confirmation models. Within a corpus the two models share data order and differ only in initialization. They are evaluated on the primary panels of the confirmation study (512 three-value cases per corpus), at its primary site (write 2 and read 3 on the first answer token) and with its metrics, and the panels and the analysis were fixed before the first evaluation. A second panel of 512 three-value cases per corpus, built from new worlds, repeats every test.

\begin{table}[!htbp]
\caption{Main models on the primary panel and on the second panel (512 three-value cases per model). Accidental $Z$: the native output equals $Z$. Changed: the write-2 transplant changes the native output. Restore: restoring the receiver's round-3 queries gives the native output. Query-only: installing only the transplant-induced queries reproduces the transplant output. Joint: both hold, among changed cases.}
\label{tab:exp-main}
\centering\scriptsize\setlength{\tabcolsep}{3pt}
\begin{tabular}{@{}llrrrrrrrr@{}}
\toprule
Panel & Model & Native $X$ & OFF $X$ & Accidental $Z$ & Transplant $Z$ & Changed & Restore & Query-only & Joint\\
\midrule
Primary & A3 & 490 & 137 & 2 & 487 & 501 & 511/512 & 512/512 & 500/501\\
 & A4 & 505 & 124 & 3 & 496 & 500 & 512/512 & 511/512 & 499/500\\
 & B3 & 503 & 119 & 1 & 496 & 505 & 512/512 & 512/512 & 505/505\\
 & B4 & 508 & 108 & 3 & 511 & 509 & 512/512 & 512/512 & 509/509\\
 & Total & 2,006 & 488 & 9 & 1,990 & 2,015 & 2,047/2,048 & 2,047/2,048 & 2,013/2,015\\
\midrule
Second & A3 & 496 & 107 & 5 & 481 & 491 & 511/512 & 512/512 & 490/491\\
 & A4 & 502 & 128 & 5 & 501 & 500 & 512/512 & 512/512 & 500/500\\
 & B3 & 492 & 126 & 4 & 482 & 500 & 512/512 & 511/512 & 499/500\\
 & B4 & 508 & 100 & 1 & 505 & 510 & 512/512 & 512/512 & 510/510\\
 & Total & 1,998 & 461 & 15 & 1,969 & 2,001 & 2,047/2,048 & 2,047/2,048 & 1,999/2,001\\
\bottomrule
\end{tabular}
\end{table}

The following counts are for the primary panel unless stated.

\textbf{Rung 1.} Local accuracy is 2,027 of 2,048 natively and 2,011 under OFF (changes of 0, 4, 12 and 0 questions per model). The composition deficit is 353 to 400 cases per model, and OFF accuracy is 119 to 147 cases below the memory-only reference (Appendix~\ref{app:protocol-ext}).

\textbf{Rung 2.} Natively the 2,048 outputs are $X$ in 2,006 cases, $Y$ in 17, $Z$ in 9 and something else in 16. After the transplant they are $Z$ in 1,990, $Y$ in 20, $X$ in 15 and something else in 23. Among the 1,994 changed cases in which the donor's actual output is neither the native output nor $Z$, the transplant copies the donor in 12 and gives $Z$ in 1,969. The transplanted recipient gives the model's own output on the hybrid world in 2,021 of 2,048 cases (508, 499, 503 and 511). Self-replacement and a transplant at the never-read write 3 leave all 2,048 outputs unchanged. Transplanting only the lexical blocks of the packet yields $Z$ in 1,741 cases, against 41 for the control block alone.

\textbf{Rung 3.} Under restoration the number of $Z$ outputs equals the native number in every model (2, 3, 1 and 3). Matched-norm query noise leaves $Z$ in 1,890 cases, which is 92.9--96.9\% of the transplant effect per model, and returns 16 of the 2,015 changed cases to native. The transplant of writes 0--2 changes 2,026 outputs. Restoring the round-2 queries returns none of them to native, and restoring the round-3 queries returns 2,025. The final read's mass on the record needed for $Z$ is 0.006--0.009 natively and 0.52--0.63 under the transplant, and its mass on the record needed for $X$ falls from 0.69--0.76 to 0.007--0.019. Restoration gives the native values and query-only installation the transplant values. On the 550 cases whose hybrid chain (the object, the person and the location that the question links) never occurs in the training pool, the transplant changes 534 outputs and restoration returns all 534 to native (526 of 527 on the second panel). Native accuracy on the 534 cases whose own chain never occurs in training is 517.

\textbf{Third-entity control.} Installing only the round-3 queries induced by a second donor that names a third person yields that person's hybrid answer in 572 of 586 cases, against 2 natively, and gives the same output as that donor's full transplant in 586 of 586 (633 of 660, against 7, and 659 of 660 on the second panel).

\subsection{A head without the direct lexical read}
\label{app:exp-completeness}
AFGW's output head adds the receiver's final lexical read to the logits directly (Eq.~\ref{eq:fullhead}), which could favor a mechanism that runs through retrieval. This section removes that term.

\textbf{Models.} For each main model (Appendix~\ref{app:exp-main}) we trained a matched model with the same seed, corpus and training whose output head has no direct lexical-read term. Panels and analysis were fixed before the first evaluation.

\textbf{Results.} The four models answer 2,035 of 2,048 composition questions on the primary panels, and 483 with the workspace off. On a separate unseen panel their composition accuracy is 97.83--100\% in every model and direction, transplanting the source's writes 0--2 yields $Z$ in 2,021 of 2,048 cases (503--509 of 512 per model), and restoration and installation at reads 1--3 both hold in 2,016 of the 2,027 changed cases (98.8--100\% per model).

\textbf{Where the content is read.} Table~\ref{tab:exp-completeness} compares the two heads under the same transplant, on the cases of the primary panels whose hybrid chain never occurs in training. A chain is the object, the person and the location that a composition question links. With the direct term, restoring the final read returns every changed case to native. Without it, restoring reads 1--3 returns 546 of 547, and the final read alone returns 487. The read spreads over rounds when the direct term is removed, and the protocol detects this shift in where content is consumed. This head is therefore tested with the source's three packets and the receiver's reads 1--3.

\begin{table}[!htbp]
\caption{Rung 3 for the two output heads under the transplant of writes 0--2, on the cases of the primary panels whose hybrid chain never occurs in training, pooled over the four models of each head. Changed: the transplant changed the native output. Restore and noise: changed cases returned to native.}
\label{tab:exp-completeness}
\centering\small
\begin{tabular}{@{}llrrr@{}}
\toprule
Head & Site & Changed & Restore & Noise\\
\midrule
with the direct term (main models) & read 3 & 541/550 & 541/541 & 1/541\\
without it & reads 1--3 & 547/550 & 546/547 & 2/547\\
without it & read 3 alone & 547/550 & 487/547 & ---\\
\bottomrule
\end{tabular}
\end{table}

\Needspace{10\baselineskip}
\subsection{Rung 4: single-packet three-receiver evaluation}
\label{app:exp-breadth}
\textbf{Design.} Two three-receiver checkpoints (seeds 106101 and 106102), a panel of 512 joint episodes from four new families fixed in advance, and donors whose maps differ from the recipient's. The three receivers hold locations, colors, and access rights. We transplant the source's write-2 packet alone (the single-packet condition), writes 0 or 1 alone, and the trajectory of writes 0--2. Because this variant computes the workspace once on the prefix and decodes each receiver from a private cache, every prefix intervention persists through the whole answer of every receiver, and the three receivers consume one shared broadcast. Write 3 is never consumed (a write-3 transplant leaves all outputs native, 512/512).

\begin{table}[!htbp]
\caption{Joint outcomes (all three receivers; 512 episodes per seed). Accidental: native joint output equals the hybrid answers. Donor copy: joint output equals the donor's actual joint output.}
\label{tab:exp-breadth}
\centering\small
\begin{tabular}{@{}lrrrrrrr@{}}
\toprule
Seed & Native & Accidental & Write 2 & Writes 0--2 & Write 0 & Write 1 & OFF\\
\midrule
106101 & 458 & 20 & 448 & 458 & 20 & 20 & 25\\
106102 & 416 & 12 & 412 & 416 & 12 & 12 & 80\\
\bottomrule
\end{tabular}
\end{table}

\FloatBarrier
\Needspace{12\baselineskip}
\textbf{Joint results.} The single write-2 packet yields all three hybrid answers jointly in 448/512 and 412/512 episodes; writes 0 or 1 alone stay at the accidental rate (Table~\ref{tab:exp-breadth}). The donor's joint output is copied in 0/512 episodes under every transplant. Under the trajectory transplant, all 512 joint outputs equal the model's own outputs on the hybrid world in each seed, errors included, while joint correctness is 458 and 416 of 512. In each receiver's donor-informative cases, donor copies number 0/156, 0/392, and 0/276 (106101) and 4/192, 0/376, and 8/300 (106102). OFF leaves 25/512 and 80/512 jointly correct; the memory-only ceiling is 256/512, and OFF outputs are constant within every ceiling group.

\begin{table}[!htbp]
\caption{Rung 3 per receiver under the write-2 packet (analysis fixed in advance). Changed: the transplant changes that receiver's output. Restore: restoring the receiver's round-3 queries gives its native output. Query-only: installing only the induced round-3 queries reproduces the transplant output. Noise: matched-norm noise gives the native output. Last column: hybrid answers within the receiver's $X/Y/Z$-distinct cases.}
\label{tab:exp-breadth-r3}
\centering\small
\begin{tabular}{@{}llrrrrr@{}}
\toprule
Seed & Receiver & Changed & Restore & Query-only & Noise & $Z$, distinct cases\\
\midrule
106101 & location & 482/512 & 482/482 & 472/482 & 0/482 & 132/160\\
106101 & color & 480/512 & 480/480 & 480/480 & 0/480 & 348/384\\
106101 & access & 476/512 & 476/476 & 476/476 & 1/476 & 260/288\\
106102 & location & 496/512 & 496/496 & 490/496 & 2/496 & 132/160\\
106102 & color & 496/512 & 496/496 & 480/496 & 0/496 & 316/384\\
106102 & access & 496/512 & 496/496 & 496/496 & 0/496 & 256/288\\
\bottomrule
\end{tabular}
\end{table}

\textbf{Rung 3 per receiver.} Read 3 feeds only the patched receiver, so a per-receiver test is exact: patching one receiver leaves its two siblings' first-token logits, cached features, lexical reads, and full outputs bitwise unchanged (512/512 for every receiver and seed). Restoration and query-only installation jointly hold in 96.8--100\% of changed cases for all three receivers in both seeds (Table~\ref{tab:exp-breadth-r3}).

These results establish breadth for a workspace that is computed once per input and read by three receivers.

\subsection{Training decides who computes: matched variants and a learned relay}
\label{app:exp-whocomputes}
\textbf{Matched variants.} For each main model we trained three variants from the same local-stage checkpoint, with the same data order and the same 12,288-update budget. In the first, the routing weights are fixed at $1/5$ on the four modules and NULL. In the second, the receiver reads both memories on composition questions, so it could answer without the workspace. In the third, the source reads both memories, so it could compute the answer and send it. All twelve models are reported (Table~\ref{tab:exp-variants}).

\begin{table}[!htbp]
\caption{Matched variants on the primary panel, pooled over four models (2,048 three-value cases). Accid.\ $Z$: the native output equals $Z$. One packet: the write-2 transplant. Writes 0--2: all three source packets. Joint: restoration and query-only installation both hold, among the outputs the one-packet transplant changes.}
\label{tab:exp-variants}
\centering\footnotesize\setlength{\tabcolsep}{3pt}
\begin{tabular}{@{}lrrrrrr@{}}
\toprule
Models & Native $X$ & OFF $X$ & Accid.\ $Z$ & One packet $Z$ & Writes 0--2 $Z$ & Joint\\
\midrule
Main models & 2,006 & 488 & 9 & 1,990 & 2,001 & 2,013/2,015\\
Uniform routing & 2,001 & 495 & 15 & 1,835 & 2,001 & 1,890/1,914\\
Receiver reads both memories & 2,002 & 458 & 14 & 1,991 & 2,001 & 2,010/2,012\\
Source reads both memories & 2,000 & 508 & 11 & 1,991 & 1,996 & 2,013/2,015\\
\bottomrule
\end{tabular}
\end{table}

\FloatBarrier
All variants match the main models in accuracy, depend on the workspace and recombine. Under uniform routing a single packet yields $Z$ less often (1,835 against 1,990 of 2,048), whereas the three-packet transplant yields $Z$ equally often, so learned routing concentrates the content in one packet (Appendix~\ref{app:exp-routing}). On the second panel the joint counts are 1,874 of 1,902 (uniform routing), 1,996 of 2,000 (receiver reads both) and 2,001 of 2,005 (source reads both).

To see whose memory the answer is computed from, we resampled one module's records at inference (three fixed draws per recipient, counting the 4,650 draws whose symbolic answer changes). Where the source reads both memories, the output follows the answer implied by the changed receiver memory in 4,578 draws. Changing the source's copy of the same facts leaves the output identical to the native one in 4,626 draws and makes it follow the changed copy in 33. Where the receiver reads both memories, the output follows its changed memory in 4,569 draws. 

\rev{\textbf{Recombination without the local stage.} The variants above start from a local-stage checkpoint (Appendix~\ref{app:training}). We also trained the two variants that read both memories from random initialization, with composition questions from the first update and the same 12,288-update budget (eight models, both initializations of each corpus). They recombine as well. They answer 500 to 512 of 512 questions on each panel, and 97 to 142 with the workspace off. Under the transplant of writes 0--2 they give the hybrid answer in 501 to 512 cases and the donor's answer in at most 7, and at inference their output follows the changed memory of the receiver in 9,253 of 9,300 draws.}

\textbf{A learned relay.} With the output head that has no direct lexical read (Appendix~\ref{app:exp-completeness}), we trained a pair of models from the same local-stage checkpoint (the first initialization of corpus A), with the same data order and budget. In both, the source reads both memories. In the relay-trained model, the receiver's records of the answer type are resampled independently on every composition training example, with the question and the answer label unchanged, so the receiver's memory carries no information about the answer and only the source can compute it. In the control, the receiver's records are intact. No packet content and no output is set to relay, and at evaluation both models see intact memories.

\begin{table}[!htbp]
\caption{Relay-trained models and their control (512 three-value cases per panel) under the transplant of the source's writes 0--2. Resampled: the share of composition training examples on which the receiver's records are resampled. Corpus, init.: the corpus and its first or second initialization. $Y$ / $Z$: outputs equal to the donor's answer and to the hybrid answer. Read tests: restoration and query-only installation at reads 1--3 both hold, among the outputs that the transplant changes.}
\label{tab:exp-learnedrelay}
\centering\footnotesize\setlength{\tabcolsep}{3.5pt}
\begin{tabular}{@{}lrllrrrr@{}}
\toprule
Model & Resampled & Corpus, init. & Panel & Native $X$ & OFF $X$ & $Y$ / $Z$ & Read tests\\
\midrule
Relay-trained & 100\% & A, 1 & primary & 511 & 45 & 511 / 0 & 0/511\\
Relay-trained & 100\% & A, 1 & second & 511 & 57 & 511 / 0 & 0/512\\
\rev{Relay-trained} & \rev{25\%} & \rev{A, 1} & \rev{primary} & \rev{512} & \rev{68} & \rev{511 / 1} & \rev{0/512}\\
\rev{Relay-trained} & \rev{25\%} & \rev{A, 1} & \rev{second} & \rev{512} & \rev{89} & \rev{512 / 0} & \rev{0/512}\\
\rev{Relay-trained} & \rev{25\%} & \rev{B, 1} & \rev{primary} & \rev{512} & \rev{77} & \rev{512 / 0} & \rev{0/512}\\
\rev{Relay-trained} & \rev{25\%} & \rev{B, 1} & \rev{second} & \rev{508} & \rev{66} & \rev{508 / 2} & \rev{0/511}\\
\rev{Relay-trained} & \rev{25\%} & \rev{A, 2} & \rev{primary} & \rev{495} & \rev{105} & \rev{482 / 18} & \rev{0/503}\\
\rev{Relay-trained} & \rev{25\%} & \rev{A, 2} & \rev{second} & \rev{500} & \rev{102} & \rev{478 / 17} & \rev{0/505}\\
\rev{Relay-trained} & \rev{25\%} & \rev{B, 2} & \rev{primary} & \rev{475} & \rev{83} & \rev{433 / 39} & \rev{9/495}\\
\rev{Relay-trained} & \rev{25\%} & \rev{B, 2} & \rev{second} & \rev{482} & \rev{68} & \rev{428 / 44} & \rev{7/498}\\
Control & 0\% & A, 1 & primary & 512 & 121 & 0 / 512 & 512/512\\
Control & 0\% & A, 1 & second & 512 & 139 & 0 / 512 & 512/512\\
\bottomrule
\end{tabular}
\end{table}

The two models cannot be told apart by accuracy, and both depend on the workspace (Table~\ref{tab:exp-learnedrelay}). The transplant gives opposite verdicts: the relay-trained model outputs the donor's answer and the control the hybrid answer. Resampling one module's records at inference confirms both verdicts (1,171 draws whose symbolic answer changes, primary panel). In the relay-trained model the output follows the changed copy of the source in 1,168 draws and the changed memory of the receiver in none. In the control it follows the changed memory of the receiver in 1,171 draws and the changed copy of the source in none. The read tests separate the two models as well. In the relay-trained model, restoring the receiver's native queries returns none of the changed outputs to native, as expected when content does not act through the receiver's retrieval.

\rev{\textbf{A standard patch does not tell the two models apart.} On a further panel of 512 held-out recipients we ran the same transplant with two kinds of donor. On this panel the relay-trained model and the control answer 512 and 510 questions, and 90 and 119 with the workspace off. A standard donor differs from the recipient only in the source's facts, so its answer is also the hybrid answer. With it, the output becomes the donor's answer in 512 of 512 cases in both models. With a three-value donor on the same recipients, the relay-trained model gives $Y$ in all 512 cases and the control gives $Z$ in all 512. The patch, its site and the models are the same in both designs. Only the choice of donor differs, and only the three-value donor shows which module computes.}

\rev{\textbf{Relay needs little pressure.} We repeated the relay training with the receiver's records resampled on a random 25\% of the composition training examples and intact on the rest, from both initializations of each corpus. All four models relay, and all fail without the workspace (Table~\ref{tab:exp-learnedrelay}). Two give the donor's answer in at least 508 of 512 cases, as the first model does. The other two give it in 428 to 482 cases and the hybrid answer in 17 to 44. At inference the output of the four follows the changed copy of the source in 4,441 of 4,650 draws and the changed memory of the receiver in 172, and a standard patch moves all four to the donor's answer (473 to 512 of 512 cases). With the receiver's memory reliable on three in four of these training examples, the source computes most or all of the answers.}
\subsection{A trained workspace model whose broadcast goes nearly unused}
\label{app:exp-earlier}
The verification gap would be empty if a trained model always used the workspace it was given. This section gives a concrete case in which it does not. The case also shows what AFGW's two boundaries are for.

\textbf{Design.} The single-reader model has the kinds of component that AFGW has: four recurrent modules, four rounds per generated token with all states reset at every token, a record memory read by learned queries, a competitive write with a NULL option, and a broadcast that every module receives in the next round and that enters the reader's queries. It differs in two respects. A single module reads all records, so a composition can be answered without any information crossing between modules. And the output head reads the workspace directly, in addition to the reader's lexical reads. The workspace has two slots, a goal slot set from the question and an evidence slot into which the modules' proposals are pooled by competition. The evidence slot is the broadcast. The model has 5,868,166 parameters.

\textbf{Task and models.} Inputs are narratives of 6 or 12 sentences in which people move between places and pick up objects (``Anna went to the bathroom.'', ``Anna picked up the milk.''). Direct questions ask where a person is. Composition questions ask where an object is, which needs its holder and then the holder's last location. We trained two models (seeds 39101 and 39102), each from a pretrained checkpoint, for 8,315 and 8,319 updates on 42,220 examples that include the 512 training families of this task. The update counts, the panels and the conditions below were fixed before training. Each model answers all 2,048 questions of each of two held-out panels of 128 families, so 8,192 of 8,192 in all. Half of the questions in each panel are compositions.

\textbf{Tests.} On the first 32 families of each panel (512 cases per model and panel, half of them compositions) we applied the two tests that need no second memory. For rung 1 we zeroed what the modules receive from the workspace in every round and at every generated token, either the broadcast alone or the broadcast and the goal slot together. For rung 2 we replaced the broadcast that the modules receive in all four rounds of the first answer token with the broadcast recorded on a donor input. The donor is the same narrative with the other person or object queried. Its answer differs from the recipient's. This tests whether the recipient's output follows the donor's answer. A further condition zeroes the workspace where the output head reads it.

\begin{table}[!htbp]
\caption{The single-reader workspace model (two models, two panels of 512 cases each). Correct: complete answer and EOS, of 1,024 cases per model. Compositions: correct composition answers, of 1,024 over both models. Unchanged: outputs identical to the native output, of 2,048 over both models.}
\label{tab:exp-earlier}
\centering\small
\begin{tabular}{@{}lrrrr@{}}
\toprule
Condition & Model 1 & Model 2 & Compositions & Unchanged\\
\midrule
No intervention & 1,024 & 1,024 & 1,024 & ---\\
Broadcast to the modules zeroed & 1,024 & 1,024 & 1,024 & 2,048\\
Broadcast and goal to the modules zeroed & 1,023 & 994 & 994 & 2,011\\
Donor broadcast to the reader, four rounds & 1,024 & 1,024 & 1,024 & 2,048\\
Donor broadcast to all modules, four rounds & 1,024 & 1,024 & 1,024 & 2,048\\
Matched-norm noise in place of the donor broadcast & 1,024 & 1,024 & 1,024 & 2,048\\
Workspace zeroed at the output head & 860 & 783 & 917 & 1,643\\
Zeroed for the modules and at the output head & 832 & 702 & 855 & 1,534\\
\bottomrule
\end{tabular}
\end{table}

\FloatBarrier
\textbf{Results.} Both models answer every case natively (Table~\ref{tab:exp-earlier}). Zeroing the broadcast changes no output. Zeroing everything the modules receive from the workspace leaves 2,017 of 2,048 answers correct, and 994 of 1,024 on compositions. A donor broadcast in all four rounds changes none of the 2,048 outputs, and no output follows the donor. The content of the broadcast therefore does not enter the computation, and nothing needs to cross between modules: the single reader chains its reads through its own recurrent state. Zeroing the workspace at the output head costs 405 answers, 298 of them on direct questions.

\textbf{Reading.} The single-reader model has every component that an architectural assessment looks for: specialized modules, a limited shared workspace filled by competition, and a broadcast to all modules. It is trained end to end and solves its task. Its broadcast is nevertheless nearly unused, and accuracy does not show this. In the terms of Section~\ref{sec:counterfactual} the protocol returns the first of its three verdicts, an ignored message. The relay-trained model of Appendix~\ref{app:exp-whocomputes} gives the second, and the main models give the third (Section~\ref{sec:content}). AFGW's two boundaries remove both ways around the broadcast. With the memory split between two readers, a composition cannot be answered inside one module. With no path from the workspace to the output, the workspace cannot serve as an output buffer. These boundaries make dependence possible, and Section~\ref{sec:necessity} shows that the trained models depend on the workspace beyond what the boundaries require. They do not decide what the broadcast carries or how the receiver uses it, which is what rungs 2 and 3 test.

\subsection{Constructed relay and recombination references}
\label{app:exp-reference}
A test of whether the three-value transplant reports relay when relay occurs needs systems whose computation is known in advance. We built two small reference designs in which the architecture fixes where the computation can happen, and trained each from random initialization. They serve as positive controls for the test and are separate from AFGW.

\textbf{Design.} Both designs read the same typed records and public question and communicate through a single 640-dimensional broadcast with a fixed schedule: the receiver writes an encoded table of its own records at write 0, and the source, which reads its own memory and this broadcast, writes a learned 32-dimensional packet at write 2. In the \emph{relay} design, the receiver's decoder sees only that packet, so the source must compute the answer and send it. In the \emph{recombination} design, the source sends information about the intermediate person, and the receiver uses the decoded person to read its own private map. No intermediate quantity is supervised. Each design was trained with two seeds (384 AdamW updates, final-answer loss only, on all training composition questions of both confirmation corpora), all four models are reported, and none was selected. Evaluation uses the unseen three-value panels of Appendix~\ref{app:exp-fresh} (1,024 composition cases per model).

\begin{table}[!htbp]
\caption{Constructed references (per model, 1,024 three-value composition cases). Accuracy is exact match on the native, donor and hybrid worlds. Transplant columns give the output class under the write-2 transplant.}
\label{tab:exp-reference}
\centering\small
\begin{tabular}{@{}llrrrrr@{}}
\toprule
Design & Seed & Accuracy (each world) & OFF correct & Transplant $=Y$ & Transplant $=Z$ & Other\\
\midrule
relay & 1 & 1,024/1,024 & 27 & 1,024 & 0 & 0\\
relay & 2 & 1,024/1,024 & 62 & 1,024 & 0 & 0\\
recombination & 1 & 1,024/1,024 & 269 & 0 & 1,024 & 0\\
recombination & 2 & 1,024/1,024 & 262 & 0 & 1,024 & 0\\
\bottomrule
\end{tabular}
\end{table}

\FloatBarrier
\textbf{Results.} Both designs solve the task perfectly, both depend on their channel (OFF leaves 27--62 and 262--269 of 1,024 correct), and in both the transplant changes every output (Table~\ref{tab:exp-reference}). Accuracy, OFF and a test of whether the transplant changes the output therefore give the same verdict on both designs. The three-value test separates them completely: the relay references output the donor's answer $Y$ in every case and the recombination references the hybrid answer $Z$ in every case. Together with Appendix~\ref{app:exp-relayarms}, which shows that the test does not report relay where trained AFGW variants recombine, this gives evidence of both sensitivity and specificity.

\subsection{A non-exclusive route: the bypass experiment}
\label{app:exp-bypass}

\textbf{Design (plan fixed in advance).} We added a learned second route between the two reader modules. At every round, a linear map of $E_2$'s state is added to $E_1$'s recurrent input at the next round, and vice versa (98,816 parameters). Each model runs the mixed stage from its lineage's local-stage checkpoint (lineages A1 and B1; 2,048 updates). In the first arm, the zero-start arm, the bypass starts at zero and is free to be learned. In the second, the forced arm, the bypass starts at a default initialization and the workspace is dropped (zero-OFF) on a random half of the training examples, so that the bypass is sometimes the only route. We analyzed each model twice. The blind analysis applies the workspace-only protocol as if the bypass were unknown, as it would be in a pretrained Transformer; the oracle analysis applies OFF and transplants to each route separately and jointly.

\begin{table}[!htbp]
\caption{The forced arm (512 three-value cases per lineage). OFF columns: correct answers ($X$). Transplant columns: cases moved to $Z$. Workspace and bypass OFF use resample ablation (zero-OFF in parentheses); the bypass-only transplant is a trajectory (writes 0--2), because a round-2 bypass message reaches the receiver only after its round-3 read.}
\label{tab:exp-bypass}
\centering\small
\begin{tabular}{@{}lrrrrrrr@{}}
\toprule
& & \multicolumn{3}{c}{OFF ($X$)} & \multicolumn{3}{c}{Transplant (moved to $Z$)}\\
\cmidrule(lr){3-5}\cmidrule(l){6-8}
Lineage & Native & workspace & bypass & both & ws, write 2 & ws, traj. & bypass, traj.\\
\midrule
A1 & 364 & 268 (342) & 240 (363) & 109 (120) & 76 & 158 & 171\\
B1 & 363 & 279 (357) & 304 (366) & 124 (147) & 134 & 211 & 123\\
\bottomrule
\end{tabular}
\end{table}

\textbf{Results.} The forced arm behaved as predicted in advance in both lineages (Table~\ref{tab:exp-bypass}). Rung 1 understates use. The bypass is used: resample-OFF of the bypass costs 124 and 59 cases, and a bypass-only transplant moves 171 and 123 cases to $Z$. Yet zero-OFF of the workspace costs only 22 and 6 cases, so a blind analysis would read the workspace as barely needed, although its content is used (below). This arm was trained with the workspace zeroed on half of its rows, so it is robust to exactly that ablation. The bound of Proposition~\ref{prop:off} therefore holds only when the workspace is the sole route. Rungs 2 and 3 survive. The workspace transplant still moves 76 and 134 cases to $Z$, against 3 and 2 under matched-norm packet noise, and restoration and query-only jointly hold in 132/136 and 173/173 changed cases. In the zero-start arm the bypass stays unused (every usage measure $\leq 5/512$), and the workspace-only verdicts are unchanged.

\textbf{Both routes are consumed through one read.} Restoring the receiver's round-3 queries also undoes the bypass-only transplant (265/266 and 154/154 changed cases), because bypass content reaches the output through the same read. Rung 3 therefore locates where content is consumed, not which route carried it. Dividing the work between routes requires per-route transplants: the trajectory effect splits 48:52 (A1) and 64:36 (B1) between workspace and bypass, nearly additively.

Resample ablation of the workspace costs 96 and 84 answers (Figure~\ref{fig:generality}a), so the kind of ablation decides what rung 1 reports here, while rungs 2 and 3 certify content use at the site with either route present (Table~\ref{tab:requirements}).

\subsection{Worlds with twice as many entities}
\label{app:exp-scale}
\textbf{Design (panels fixed in advance).} We applied rungs 1--3, without any fine-tuning, to models trained only on four-entity worlds, on worlds with eight objects, eight people and eight locations (sixteen records). New eight-entity families were generated with the original data generator. From each of the first eight confirmation families we took the first 16 legal three-value pairs per composition direction (256 composition cases per model) and four cases per local task (128 local cases), choosing by symbolic inputs only. Decoding, the write-2 transplant, round-3 restoration and query-only installation are exactly as in Section~\ref{sec:mechanism}.

\textbf{Main models.} The four main models (Appendix~\ref{app:exp-main}) answer 973 of 1,024 eight-entity composition questions, against 119 under OFF, while local accuracy changes from 496 to 487 of 512. The write-2 transplant yields $Z$ in 973 of 1,024 cases, against 4 natively, and changes 1,008 outputs. Restoration returns 1,022 of 1,024 outputs to native, query-only installation reproduces the transplant in 1,023, and both hold in 1,005 of the 1,008 changed cases. The mechanism identified on four-entity worlds thus carries over, without retraining, to worlds twice as large.

\subsection{Other communication architectures}
\label{app:exp-otherarch}
We applied rungs 2 and 3 to two variants: \emph{p2p}, in which each receiver attends over the other modules' packets, and \emph{slot}, a Goyal-style shared workspace of $4\times160$ slots \citep{goyal2022}. Each has two models (seeds 104101 and 104102), trained on a separate corpus of eight families, and is evaluated on the unseen corpus-A panel (512 three-value cases). For p2p, the transplant yields $Z$ in 410 and 399 cases (accidental 23 and 18), and restoration and query-only jointly hold in 454/454 and 444/447 changed cases. For the slot workspace, the transplant changes fewer outputs (286 and 280) and yields $Z$ in 235 and 213 cases (accidental 22 and 24), with the joint rung-3 equalities in 277/286 and 264/280 changed cases; matched-norm noise returns 36/286 and 30/280 of them to native. \rev{The slot workspace spreads its content over its writes. Transplanting the source's writes 0--2 yields $Z$ in 432 and 416 cases, as often as the models answer the hybrid world natively (433 and 416), and restoration and installation at reads 1--3 then jointly hold in 469/471 and 457/462 changed cases.}

\textbf{An order-invariant reader.} AFGW's reader favors the last matching record (Appendix~\ref{app:design}), so reordering identical facts can change what it reads; this explains the 13--32 of 128 outputs that changed in the record-order control. To check that the mechanism does not depend on this order sensitivity, we evaluated a variant with a permutation-invariant set reader (Appendix~\ref{app:analysis-order}; seeds 104101 and 104102) on the same unseen corpus-A panel. Reordering the records changes none of its native, donor or transplant outputs (0/512 each). Its signature is intact: the transplant yields $Z$ in 392 and 426 cases (accidental 35 and 24), and restoration and query-only jointly hold in 424/424 and 442/442 changed cases.

\Needspace{30\baselineskip}
\subsection{Rungs 2 and 3 at a known carrier in a pretrained Transformer}
\label{app:exp-qwen}
This experiment runs rungs 2 and 3 where the channel is not exclusive and the read interface is attention. It tests the portability of the protocol to a pretrained Transformer, with a carrier that is known by construction.

\textbf{Design.} We used Qwen2.5-1.5B-Instruct with greedy decoding in a two-stage task with an explicit carrier. In the first stage the model answers an ownership question with a name; the generated name token is the carrier. In the second stage a new user turn shows a register that maps names to locker codes and asks for the code of the name in the previous answer. The native world gives answer $X$ (the code of the native owner), the donor world gives $Y$, the hybrid world (the donor's ownership, the native register) gives $Z$, and $X$, $Y$ and $Z$ are pairwise distinct. The transplant replaces the carrier's layer-4 residual with the donor's. For rung 3 we restore or install the answer position's post-softmax attention rows over a read window, with values left live. The window, layers 20--27, was chosen by a rule applied to native attention on a development panel alone (the layers that hold 80\% of the excess attention to the owner's register row). The rule, the data and the analysis were fixed before the evaluation panel was run, an unseen panel of 128 cases from 16 families.

\textbf{Results.} The transplant yields $Z$ in 71 of 128 cases, against 12 natively (59 gains, no losses; $Z$ exceeds the native rate in 16/16 families), and it copies the donor's own answer in 11, against 5 natively. It changes the native output in 72 cases. On those, restoring the native attention rows at the window returns the native output in 71/72, installing the transplant-induced rows into the native run reproduces the transplant output in 71/72, and both hold in 70/72. Random rows matched to the same per-head total-variation distance give 12/72 and 9/72 (joint 2/72), and an equally sized early window (layers 5--12) gives 0/72 throughout. Restoring the values alone, with transplant rows, returns only 18/72, so the effect runs through where the answer position reads rather than what it reads. Mean-ablating the carrier's keys and values at all layers lowers native accuracy from 100/128 to 69/128, so the carrier is not an exclusive route, and an OFF test would understate its use.

\end{document}